\documentclass[lettersize, journal]{IEEEtran} 
\IEEEoverridecommandlockouts                              
\usepackage{xcolor}
\usepackage{amsmath}
\usepackage{amssymb}
\usepackage{dsfont}
\usepackage{graphicx}
\usepackage{algorithm}
\usepackage{algpseudocode}
\usepackage{romannum}
\usepackage{hyperref}
\usepackage[export]{adjustbox}
\usepackage{subcaption}
\usepackage{pdfpages}
\usepackage{enumitem}
\usepackage{soul}
\usepackage{tikz}
\DeclareFontFamily{OT1}{pzc}{}
\DeclareFontShape{OT1}{pzc}{m}{it}{<-> s * [1.10] pzcmi7t}{}
\DeclareMathAlphabet{\mathpzc}{OT1}{pzc}{m}{it}

\newcommand\copyrighttext{%
  \footnotesize \textcopyright~ 2024 IEEE. Personal use of this material is permitted.  Permission from IEEE must be obtained for all other uses, in any current or future media, including reprinting/republishing this material for advertising or promotional purposes, creating new collective works, for resale or redistribution to servers or lists, or reuse of any copyrighted component of this work in other works.}
\newcommand\copyrightnotice{%
\begin{tikzpicture}[remember picture,overlay]
\node[anchor=south,yshift=5pt] at (current page.south) {\fbox{\parbox{\dimexpr\textwidth-\fboxsep-\fboxrule\relax}{\copyrighttext}}};
\end{tikzpicture}%
}

\graphicspath{{./figs/}{./figs/sim/}{./figs/real/}{./figs/data/}{./figs/adapt/}}

\title{
Attribute-Based Robotic Grasping with Data-Efficient Adaptation
}

\author{Yang Yang, Houjian Yu, Xibai Lou, Yuanhao Liu, and Changhyun Choi
\thanks{Manuscript received 8 August 2023; accepted 22 December 2023. Date of publication 12 January 2024; date of current version 31 January 2024. This paper was recommended for publication by Associate Editor M. Walter and Editor J. Bohg upon evaluation of the reviewers’ comments. This work was supported in part by the NSF under Grant 2143730, in part by the Sony Research Award Program, UMII-MnDRIVE Ph.D. Graduate Assistantship, and MnDRIVE Initiative on Robotics, Sensors, and Advanced Manufacturing [DOI: 10.1109/ICRA48506.2021.9561139]. This paper is an evolved version of the conference paper \cite{yang2021attribute}. (Yang Yang and Houjian Yu are joint first authors.) (Corresponding author: Yang Yang.)}
\thanks{The authors are with the University of Minnesota, Minneapolis, MN 55413 USA (e-mail: yang5276@umn.edu; yu000487@umn.edu; lou00015@umn.edu; liu00800@umn.edu; cchoi@umn.edu).}%
\thanks{Supplementary material is available at https://z.umn.edu/attr-grasp.}
\thanks{This article has supplementary downloadable material available at https://doi.org/10.1109/TRO.2024.3353484, provided by the authors.}
\thanks{Digital Object Identifier 10.1109/TRO.2024.3353484}
}

\begin{document}

\maketitle

\copyrightnotice

\begin{abstract}
Robotic grasping is one of the most fundamental robotic manipulation tasks and has been the subject of extensive research. However, swiftly teaching a robot to grasp a novel target object in clutter remains challenging. This paper attempts to address the challenge by leveraging object attributes that facilitate recognition, grasping, and rapid adaptation to new domains. In this work, we present an end-to-end encoder-decoder network to learn attribute-based robotic grasping with data-efficient adaptation capability. We first pre-train the end-to-end model with a variety of basic objects to learn generic attribute representation for recognition and grasping. Our approach fuses the embeddings of a workspace image and a query text using a gated-attention mechanism and learns to predict instance grasping affordances. To train the joint embedding space of visual and textual attributes, the robot utilizes object persistence before and after grasping. Our model is self-supervised in a simulation that only uses basic objects of various colors and shapes but generalizes to novel objects in new environments. To further facilitate generalization, we propose two adaptation methods, adversarial adaption and one-grasp adaptation. Adversarial adaptation regulates the image encoder using augmented data of unlabeled images, whereas one-grasp adaptation updates the overall end-to-end model using augmented data from one grasp trial. Both adaptation methods are data-efficient and considerably improve instance grasping performance. Experimental results in both simulation and the real world demonstrate that our approach achieves over 81\% instance grasping success rate on unknown objects, which outperforms several baselines by large margins.
\end{abstract}

\begin{IEEEkeywords}
Grasping, Deep Learning in Grasping and Manipulation, Perception for Grasping and Manipulation
\end{IEEEkeywords}

\section{Introduction}
\begin{figure}[!t]
  \begin{subfigure}{0.24\textwidth}
    \includegraphics[width=\textwidth]{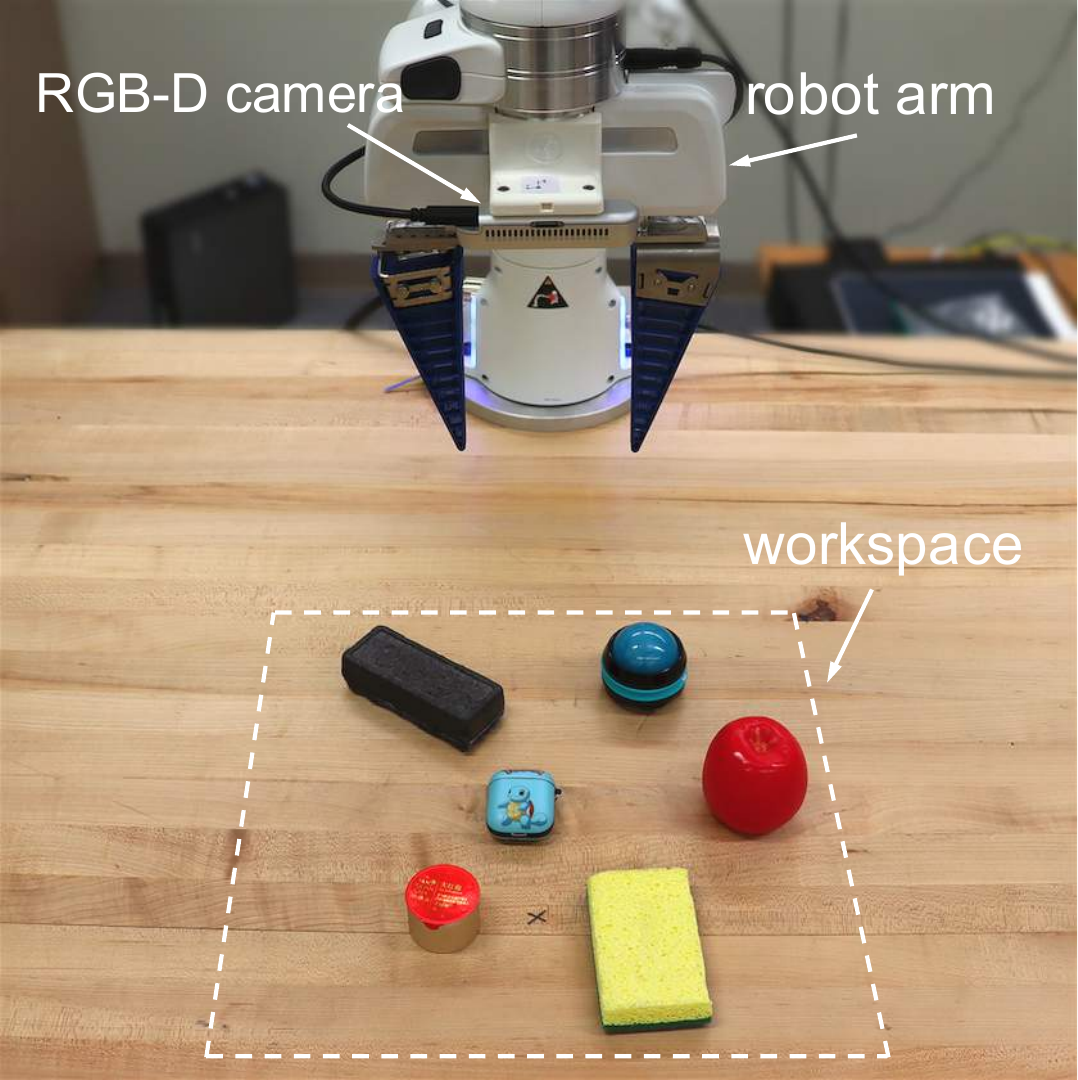}
    \caption{Objects of various attributes}
    \label{fig:intro_a}
  \end{subfigure}
  \hfill
  \begin{subfigure}{0.24\textwidth}
    \includegraphics[width=\textwidth]{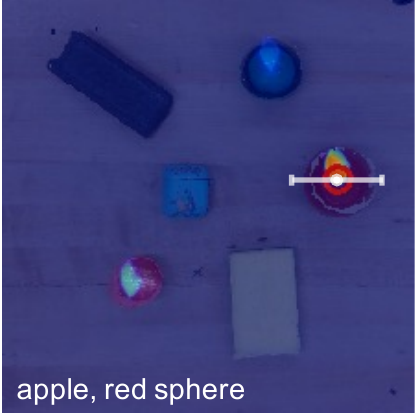}
    \caption{Grasping affordances}
  \end{subfigure}
  \caption{\textbf{Attribute-based instance grasping}. Various objects of generic attributes are placed in the workspace, and we propose to grasp a target object by describing its attributes, e.g., ``Please give me the apple, a red sphere.''.}
  \label{fig:intro}
\end{figure}

\IEEEPARstart{O}{bject} attributes are generalizable properties in object manipulation. Imagine how we describe a novel object when asking someone to fetch it, ``Please give me the apple, a red sphere.'', we intuitively characterize the target by its appearance attributes (see Fig. \ref{fig:intro}). If an assistive robot can be similarly commanded utilizing such object attributes (e.g., color, shape, and category name, etc.), it would allow better generalization capability for novel objects than using a discrete set of pre-defined category labels. Moreover, individuals learn to recognize and grasp an unknown object through rapid interactions; hence, it would be advantageous if a grasping pipeline is capable of adapting with minimal adaptation data. These factors motivate the development of attribute-based robotic grasping with data-efficient adaptation capability.

Recognizing and grasping a target object in clutter is crucial for an autonomous robot to perform daily-life tasks in the real world. Over the past years, the robotics community has made substantial progress in target-driven robotic grasping by combining off-the-shelf object recognition modules with data-driven grasping models \cite{fang2018multi}, \cite{yang2020deep}. However, these recognition-based approaches presume a unique ID for each category and are likely to experience limited generalization when applied to novel objects. In contrast, we propose an attribute-based robotic grasping approach that enables a robot to grasp an attributes-specified target object. The intuition of using attributes for grasping is that the grounded attributes can help transfer object recognition and grasping capabilities across different environments.

Suffering from domain shift \cite{quinonero2008covariate}, a machine learning model trained with the data in one domain is subject to limited generalization when tested in another domain. In robotic grasping, the source of domain shifts includes novel objects, new environments, perception noises, etc. To mitigate the domain shift, domain adaptation methods \cite{ben2007analysis} are widely used for model transfer. These adaptation methods, on the other hand, typically require the collection of a large adaption dataset, which is costly, inefficient, and time-consuming. To efficiently transfer our pre-trained attribute-based grasping model, we present two tailored adaptation methods. Both the two proposed adaptation methods are data-efficient, requiring minimal data collecting and labeling.

Compared to recognition-based robotic grasping (i.e., employing pre-trained recognition modules), the challenges of attribute-based grasping are 1) mapping from workspace images and query text of the target to robot motions, 2) associating abstract attributes with raw pixels, 3) data labeling in target-driven grasping, and 4) data-efficient adaptation to unknown objects and new scenes. In this paper, we design an architecture that consists of a multimodal encoder (i.e., encoding both visual and textual data) and an affordances decoder (i.e., predicting instance grasping affordances \cite{zeng2019learning}). The key aspects of our system are:
\begin{itemize}
    \item We design the deep grasping neural networks that represent 3-DOF grasp poses. After encoding and fusing visual-textual representations, the networks rotate the fused features to account for different grasping angles, and then predict pixel-wise instance grasping affordances.
    \item To learn a multimodal metric space, we employ the equation of object persistence before and after grasping; the visual embedding of a grasped object should be equal to the textual embedding of that object.
    \item Our model learns object attributes that generalize to new objects and scenes by only using basic objects (of various colors and shapes) in simulation.
    \item With the pre-trained attribute representations, our model supports efficient adaptation with minimal data. Adversarial adaptation regulates the image encoder with augmented data of unlabeled images, whereas one-grasp adaptation updates the end-to-end model with augmented data requiring only one successful grasp trial. Both adaption approaches are data-efficient, and they can be employed independently or in combination to improve instance grasping performance. 
\end{itemize}
The deep grasping model in our approach is fully self-supervised through the interactions between the robot and objects. Fig. \ref{fig:intro} presents an example of attribute-based robotic grasping, wherein our approach successfully grounds object attributes and accurately predicts grasping affordances for an attributes-specified target object.

In our prior work \cite{yang2021attribute}, we proposed 1) an end-to-end architecture for learning text-commanded robotic manipulation and 2) a method of self-supervising multimodal attribute embeddings through object grasping to facilitate quick adaptation. As an evolved paper, this article presents an in-depth study of adaptation in robotic manipulation and strives to improve the autonomy of robots by achieving self-supervision and self-adaptation. The pre-trained model is self-supervised in a simulation that only uses basic objects of various colors and shapes. In our adaptation framework, we make use of autonomous robots to collect raw data for adaptation. We present three core technical contributions as follows: 
\begin{itemize}[noitemsep, topsep=0pt]
    \item[1)] \textbf{A sequential adaptation scheme.} We propose a robotic grasping adaptation framework that comprises two stackable and data-efficient adaptation methods. The adversarial adaptation and one-grasp adaptation methods aim to comprehensively adapt the model for object recognition and grasping. Through data-efficient adaptation, the robot adeptly grasps challenging objects, eliminating the need for extensive data collection.
    \item[2)] \textbf{Data-efficient augmentation methods.} We design data augmentation methods that only require unlabeled images of candidate objects for adversarial adaptation and one-grasp data of a target object for one-grasp adaptation.
    \item[3)] \textbf{Evaluation and analysis of robot grasping.} We evaluate the grasping model in simulated and real-world scenes with various testing objects and domain gaps, which verifies the effectiveness of our grasping model. Furthermore, the ablative analysis of the data augmentation methods shows the efficiency of our approach.
\end{itemize}
With observations from an RGB-D camera, our robot system is designed to grasp a target object following the user command containing object attributes.
To our best knowledge, this is the first work that explores object attributes to improve the generalization and adaptation of deep robotic grasping models. 
We believe that the adaptation framework not only enhances the overall performance but also opens up new possibilities for solving the problem in target-driven robotic manipulation.

\section{Related Work}
\subsection{Instance Grasping}
Though there are different taxonomies, the existing work of robotic grasping can be roughly divided according to approaches and tasks: 1) model-driven \cite{sahbani2012overview} and data-driven \cite{bohg2013data} approaches; 2) indiscriminate \cite{lou2020learning} \cite{zeng2018learning} and instance grasping \cite{fang2018multi} tasks. Our approach is data-driven and focuses on instance grasping. Typical instance grasping pipelines assume a pre-trained object recognition module (e.g., detection \cite{fang2018multi}, segmentation \cite{yang2020deep} \cite{yu2022self}, template matching \cite{danielczuk2019mechanical}, and object representation \cite{jang2018grasp2vec}, etc.), limiting the generalization for unknown objects and the scalability of grasping pipelines. Our model is end-to-end and exploits object attributes for generalization. Some recent research also proposes end-to-end learning methods for instance robotic grasping. \cite{jang2017end} learns to predict the grasp configuration for an object class with a learning framework composed of object detection, classification, and grasp planning. In \cite{cai2020ccan}, CCAN, an attention-based network, learns to locate the target object and predict the corresponding grasp affordances given a query image. Compared to these methods, the main features of our work are two-fold. First, we collect a much smaller dataset of synthetic basic objects to learn generic attribute-based grasping. Moreover, our generic grasping model is capable of further adapting to new objects and domains. Second, our approach takes a description text of target attributes as a query command, which is more flexible when grasping a novel object.

\subsection{Attribute-Based Methods}
Object attributes are middle-level abstractions of object properties and generalizable across object categories \cite{farhadi2009describing}. Learning object attributes has been widely studied in the tasks of object recognition \cite{zhong2000object}, \cite{sun2013attribute}, \cite{hermans2011affordance}, \cite{pirk2020online}, while attribute-based robotic grasping has been much less explored, except for \cite{cohen2019grounding}, \cite{ahn2018interactive}. Cohen et al. \cite{cohen2019grounding} developed a robotic system to pick up the target object corresponding to a description of attributes. Their approach minimizes the cosine similarity loss between visual and textual embeddings as well as predicts object attributes. However, they only show generalization across viewpoints but not object categories. In \cite{ahn2018interactive}, the proposed Text2Pickup system uses object attributes to specify a target object and removes ambiguities in the user’s command. They use mono-color blocks as training and testing objects but fail to show generalization to novel objects. In contrast, our work learns generic attribute-based robotic grasping (only using synthetic basic objects) and generalizes well to novel objects and real-world scenes.

\subsection{Model Generalization}
Model generalization is one of the most important challenges in robotic manipulation. To improve model generalization, various approaches to bridging domain gaps have been proposed. Domain randomization \cite{sadeghi2016cad2rl} is one frequently used method, which collects more diverse data by randomizing simulation settings. Some recent research \cite{tobin2017domain}, \cite{tobin2018domain}, \cite{mehta2020active} have applied domain randomization to improve the real-world generalization of a simulation-trained robot policy. We build a simulation environment and apply domain randomization during the pre-training of a generic model. In addition to domain randomization, we propose two adaption methods following the form of domain adaptation and few-shot learning. Domain adaptation \cite{ben2007analysis}, a subcategory of transfer learning \cite{pan2009survey}, is used to reduce the domain shift between the source and target domain when the feature space is the same but the distributions are different. Inspired by adversarial domain adaptation \cite{tzeng2017adversarial}, our approach learns a domain classifier and the image encoder learns domain-invariant features to confuse the classifier. We propose an object-level augmentation method to enrich the image dataset for adversarial training, increasing the generalization of the encoder to new domains. While there exist similar work, for example, Chen et al. \cite{Chen2023} investigated domain adversarial training in their work, their approach focuses on updating the feature adaptor and the discriminator using unlabeled data, rather than updating the grasp synthesis model. In contrast, our grasping adaptation approach, consisting of unsupervised adversarial adaptation and supervised few-shot learning, jointly updates the grasping pipeline.

Few-shot learning \cite{fei2006one} is the paradigm of learning from a small number of examples at test time. The key of metric-based few-shot learning method, one of the most popular categories, is to supervise the latent space and learn a versatile similarity function by metric loss \cite{koch2015siamese}, \cite{snell2017prototypical}. The supervised metric space supports fine-tuning using minimal adaptation data (also known as support set), and the similarity function generalizes to unknown test data \cite{motiian2017few}, \cite{dhillon2019baseline}. Motivated by the idea of few-shot learning methods, our approach first learns a joint metric space that encodes object attributes and then fine-tunes recognition and grasping of our model when testing on novel objects.

\begin{figure*}[!t]
  \centering
  \includegraphics[width=\textwidth]{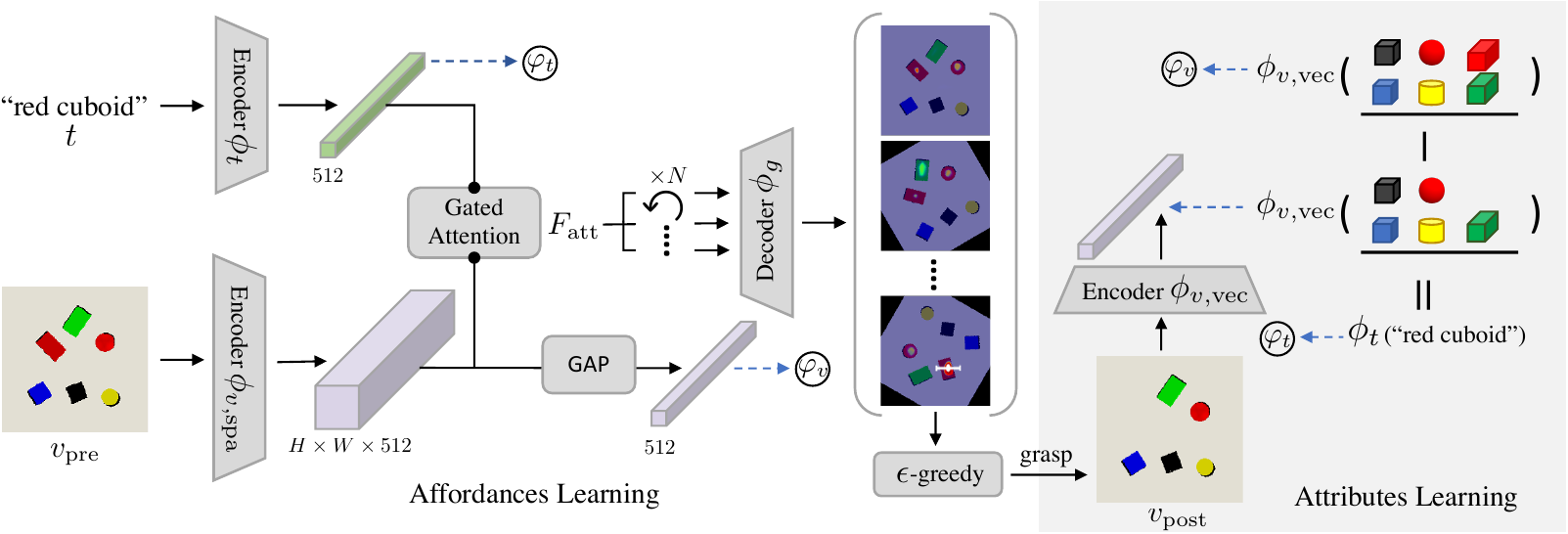}
  \vspace{2pt}
  \caption{\textbf{Overview of affordances and attribute learning.} The workspace image and query text are encoded separately and fused using gated-attention. The fusion matrix $F_{\text{att}}$ is rotated by $N$ orientations for different grasping angles and then fed into the grasping affordances decoder. The decoder learns to predict pixel-wise scores of target grasping success, and we run the $\epsilon$-greedy grasping policy and obtain the image $v_{\text{post}}$ after grasping. By utilizing the equation of object persistence before and after grasping, we learn a metric space where multimodal embedding vectors corresponding to similar attributes are encouraged to be closer. Note we denote the combination of $\phi_{v, \text{spa}}$ and GAP as $\phi_{v, \text{vec}}$, which encodes images to vectors.}
  \label{fig:overview}
\end{figure*}
%

\section{Problem Formulation}
The attribute-based robotic grasping problem in this paper is formulated as follows:

\vspace{5pt}\noindent\textbf{Definition 1.} \textit{Given a query text for a target object, the goal for the robot is to grasp the corresponding object that is placed in the cluttered workspace.}\vspace{5pt}

To handle the natural language that is diverse and unconstrained, we assume a language attribute parser, such as \cite{kazemzadeh2014referitgame}, and make the following assumption:

\vspace{5pt}\noindent\textbf{Assumption 1.} \textit{The query text is parsed into the keywords of object attributes as an input to the robotic grasping model.}\vspace{5pt}

We consider color, shape, and category name attributes in this paper, while the proposed approach is extensible to other attributes (e.g., texture, pose, and functionality, etc.). In order to make object recognition tractable, we have the following assumption regarding object placement:

\vspace{5pt}\noindent\textbf{Assumption 2.} \textit{The objects are stably placed within the workspace, and there is no stacking between objects.}\vspace{5pt}

While we show robotic grasping as a manipulation example in this paper, the proposed attribute-based learning methods should be, in principle, extensible to other robotic manipulation skills, such as suction, pushing, and placing.

\section{Learning Attribute-Based Grasping}
Object attributes are semantically meaningful features and serve as an intermediate representation for object recognition and manipulation. In this section, we propose an end-to-end neural network for attribute-based robotic grasping. The proposed model takes as input an image of visual observation and a text of target description to predict pixel-wise instance grasping affordances. To acquire a rich dataset for training, we build a simulation environment that allows domain randomization with diverse objects. In simulation, the model is pre-trained to learn instance grasping and object attributes simultaneously.

\subsection{Learning Grasping Affordances}
\label{subsec:method_grasp}
We formulate attribute-based grasping as a mapping from pairs of workspace images and query text to target grasping affordances. The proposed visual-textual manipulation architecture assumes no prior linguistic or perceptual knowledge. It consists of two modules, a multimodal encoder and an affordances decoder, as illustrated in Fig. \ref{fig:overview}.

\textbf{Multimodal Encoder}:
As shown in Fig. \ref{fig:intro_a}, our robot system uses an overhead RGB-D camera to capture the workspace. The RGB-D image is projected into a 3-D point cloud and then orthographically back-projected in the gravity direction to construct a heightmap image $v_{\text{pre}}$ of RGB and depth channel. To specify an object in the image as the grasping target, we give a text command $t$ composed of color and/or shape attributes, e.g., ``red cuboid''. The workspace image $v_{\text{pre}}$ and query text $t$ are the input to visual spatial encoder $\phi_{v, \text{spa}}$ and text encoder $\phi_t$ respectively. We use the ImageNet-pretrained \cite{deng2009imagenet} ResNet-18 \cite{he2016deep} backbone as our image encoder $\phi_{v, \text{spa}}$. We replace the first convolutional layer of the ResNet backbone with a 4-channel convolutional layer to match the RGB-D heightmap input. The encoder encodes the RGB and depth observation into 3D visual matrix $\varphi_{v, \text{spa}} \in \mathbb{R}^{H \times W\times 512}$. The text encoder $\phi_t$ is a deep averaging network \cite{iyyer2015deep} represented by three fully-connected layers and interleaved ReLU \cite{nair2010rectified} activation functions. We first map each token in a sentence text to an embeddings vector of $128$ dimension. The mean token embeddings (i.e., continuous bag-of-words \cite{mikolov2013efficient} model) of the text are input to the 3-layer MLP text encoder to produce a text vector $\varphi_{t} \in \mathbb{R}^{512}$. The visual matrix $\varphi_{v, \text{spa}}$ and the text vector $\varphi_{t}$ are then fused by the gated-attention mechanism \cite{chaplot2018gated}: each element of $\varphi_{t}$ is repeated and expanded to an $H \times W$ matrix to match the dimension of $\varphi_{v, \text{spa}}$. The expanded matrix is multiplied element-wise with $\varphi_{v, \text{spa}}$ to produce a fusion matrix $F_\text{att}$. The gated-attention unit is designed to gate certain pixels in the visual feature matrix matching to the text vector, resulting in the fusion matrix containing the visual features selected by the query text. By this means, we can detect different attributes of the objects in the image, such as color and shape.

\textbf{Affordances Decoder}:
Grasping affordances decoder $\phi_g$ is a fully-convolutional residual network \cite{he2016deep}, \cite{long2015fully} interleaved with spatial bilinear $4\times$ upsampling and ended with the sigmoid function. The decoder takes as input the fusion matrix $F_\text{att}$ and outputs a unit-ranged map $Q_g$ with the same size and resolution as the input image $v_{\text{pre}}$. Each value of a pixel $q_i \in Q_g$ represents the predicted score of target grasping success when executing a top-down grasp at the corresponding 3D location with a parallel-jaw gripper oriented horizontally concerning the map $Q_g$. The grasping primitive is parameterized by a 3-D location and an angle. To examine different grasping angles, we rotate the input $F_\text{att}$ by $N = 6$ (multiples of $30^{\circ}$) orientations before feeding into the decoder, which predicts pixel-wise scores of horizontal grasps within the rotated heightmaps. The pixel with the highest score among all the $N$ maps determines the parameters (i.e., location and angle) for the grasping primitive to be executed. As in Fig. \ref{fig:overview}, our model predicts accurate target grasping location and valid (e.g., the selected angles for the red cuboid) target grasping angle.

The motion loss $\mathcal{L}_{grasp}$, which supervises the entire encoder-decoder networks, is the error from predictions of grasping affordances:
\begin{align}
    \label{eq:motion}
    \mathcal{L}_{grasp} = \sum^{N_\mathpzc{s}} \left[(q_e-\bar{q}_e)^2 + \lambda_M \sum_{i \in M} q_i^2 \right]
\end{align}
where $N_s$ is the size of the dataset that is collected in simulation, $q_e$ is the grasping score in $Q_g$ at the executed location, and $\bar{q}_e$ is the ground-truth label (see Sec. \ref{subsec:method_data}). The second term ensures lower grasping scores for the pixels in background mask $M$ (obtained from the depth image) with weight $\lambda_M$ \cite{liang2019knowledge}, and $q_i$ is the grasping score of a background pixel.
\subsection{Learning Multimodal Attributes} \label{subsec:method_metric}
\begin{figure}[!t]
  \centering
  \includegraphics[width=0.48\textwidth]{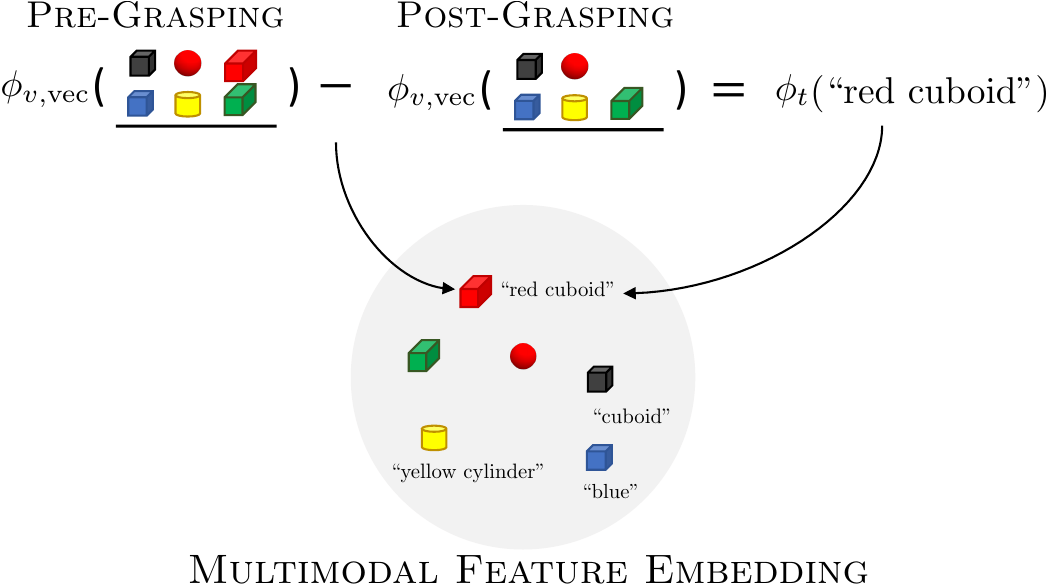}
  \caption{\textbf{Multimodal feature space} supervised by the equation of object persistence (\ref{eq:metric}). The image and text encoders are trained to produce consistent embeddings, where feature vectors corresponding to similar attributes are encouraged to be closer.}
  \label{fig:metric_feat}
\end{figure}
To learn generic object attributes, we perform multimodal attributes learning, where visual or textual embedding vectors corresponding to similar attributes are encouraged to be closer in the latent space. Inspired by \cite{jang2018grasp2vec}, we take advantage of the object persistence: the embedding difference of the scene before and after grasping is enforced closer to the representation of the grasped object. During data collection, we record image-text data $(v_\text{pre}, v_\text{post}, t)$, where $v_\text{pre}$ and $v_\text{post}$ are the workspace image before and after grasping respectively, and $t$ is the query text that describes attributes of the grasped object.

We add one layer of global average pooling (GAP) \cite{lin2013network}, \cite{zhou2016learning} at the end of the encoder $\phi_{v, \text{spa}}$ and denote the network as visual vector encoder $\phi_{v, \text{vec}}$. The output from $\phi_{v, \text{vec}}$ is a visual embedding vector that represents the average of scene features and has the same dimension of $\phi_t(t)$. We express the logic of the object persistence as an arithmetic constraint on visual and textual vectors such that $(\phi_{v, \text{vec}}(v_\text{pre}) - \phi_{v, \text{vec}}(v_\text{post}))$ is close to $\phi_t(t)$. We use the triplet loss \cite{schroff2015facenet} to approximate the constraint, and the set of triplets $\mathcal{T}$ is defined as
\begin{align}
    \mathcal{T} = \left\{(f_i, f_i^+, f_i^-) \mid s(\mathpzc{a}_{f_i}, \mathpzc{a}_{f_i^+}) > s(\mathpzc{a}_{f_i}, \mathpzc{a}_{f_i^-})\right\}
\end{align}
where $f_i$, $f_i^+$ and $f_i^-$ are random samples from the pool of vectors $(\phi_{v, \text{vec}}(v_\text{pre}) - \phi_{v, \text{vec}}(v_\text{post}))$ and $\phi_t(t)$, and $\mathpzc{a}_{f}$ is an $n$-dimensional attribute label vector corresponding to the feature vector $f$ (e.g., color, shape, and category name, etc.). 
Function $s(\cdot, \cdot)$ is an attribute similarity function that evaluates the similarity between two attribute label vectors:
\begin{align}
    \label{eq:similarity}
    s(\mathpzc{a}_1, \mathpzc{a}_2) &= \frac{1}{n}\sum_{i=1}^n \mathds{1}(\mathpzc{a}_1^i, \mathpzc{a}_2^i)\\
    \mathds{1}(\mathpzc{a}_1^i, \mathpzc{a}_2^i) &= \begin{cases} 
      1 & \text{if}~\mathpzc{a}_1^i = \mathpzc{a}_2^i \neq 0 \\
      0 & \text{otherwise}
    \end{cases}
\end{align}
where $\mathpzc{a}^i$ denotes the $i$-th element of the label vector $\mathpzc{a}$, and the indicator function $\mathds{1}(\cdot, \cdot)$ evaluates the element-wise similarity. Note that $0$ indicates null attribute meaning no attribute is specified in the label. As an example, suppose we have the dictionary $\text{dict}=\{``eos": 0, ``red":1, ``black":2, ``yellow":3, ``cylinder":4, ``cube":5\}$; then, ``red cylinder" can be represented as a label vector $\mathpzc{a}_{f_0}=[1,4]$, and ``red cube" can be represented as $\mathpzc{a}_{f_1}=[1,5]$. The similarity between the two label vectors is computed using (\ref{eq:similarity}) such that $s(\mathpzc{a}_{f_0}, \mathpzc{a}_{f_1})=s([1,4], [1,5])=0.5$. 
Additionally, when ``red'' and ``black'' are used without any additional attribute description, they are mapped to the vectors $[1, 0]$ and $[2, 0]$, respectively. In this case, the similarity between the two labels is derived as $s([1, 0], [2, 0]) = 0$. With the triplets of embedding vectors, multimodal metric loss $\mathcal{L}_{attr}$ is defined as
\begin{align}
    \label{eq:metric}
    \mathcal{L}_{attr}(\mathcal{T})= \sum_{i=1}^{|\mathcal{T}|} \max \left(\|f_i - f_i^+\|^2-\|f_i- f_i^-\|^2 + \alpha, 0\right)
\end{align}
where $\alpha$ is a hyperparameter that controls the margin between positive and negative pairs. By encoding workspace images and query text into a joint metric space and supervising the embeddings through the equation of object persistence (as shown in Fig. \ref{fig:metric_feat}), we learn generic attributes that are consistent across object categories, as discussed in Sec. \ref{subsec:exp_metric}. 
\subsection{Data Collection and Training} 
\label{subsec:method_data}
\begin{figure}[!t]
  \begin{subfigure}{0.23\textwidth}
    \includegraphics[width=\textwidth]{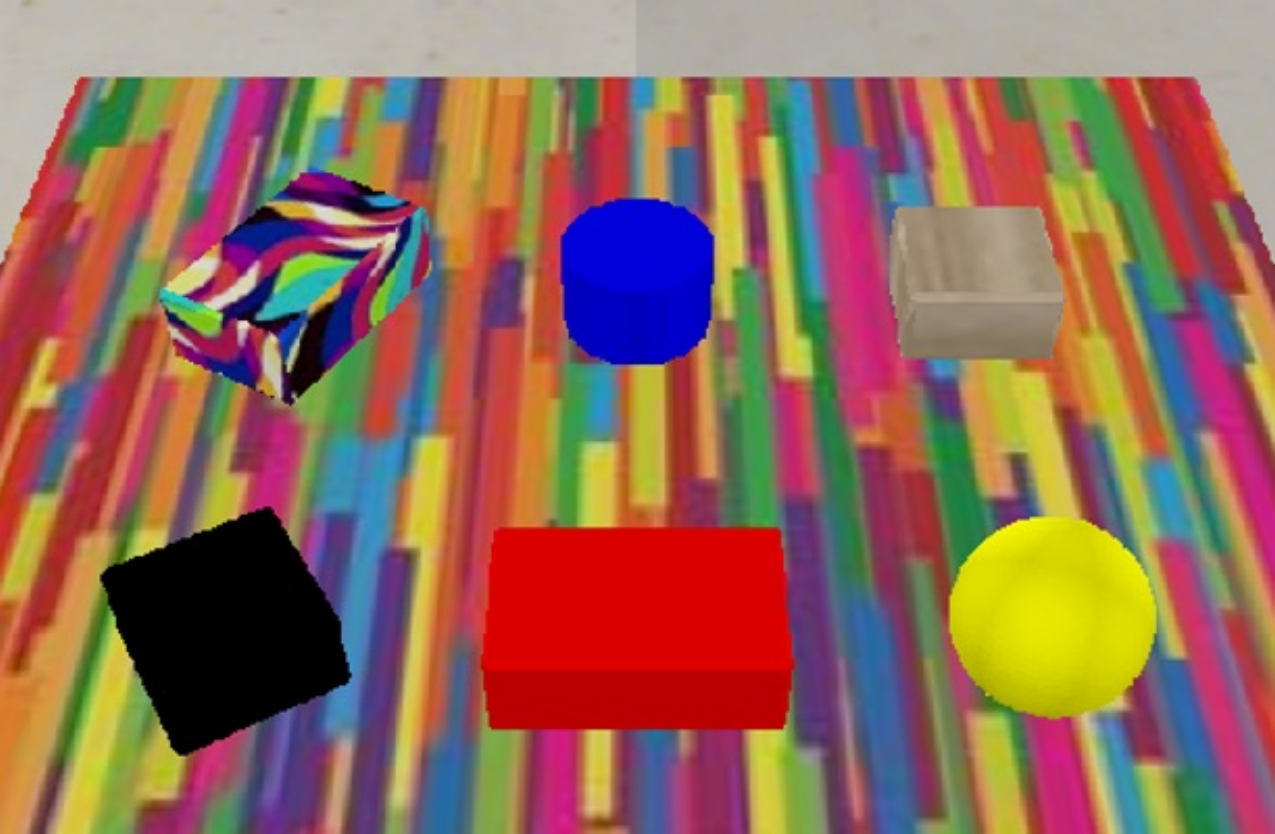}
  \end{subfigure}
  \hfill
  \begin{subfigure}{0.23\textwidth}
    \includegraphics[width=\textwidth]{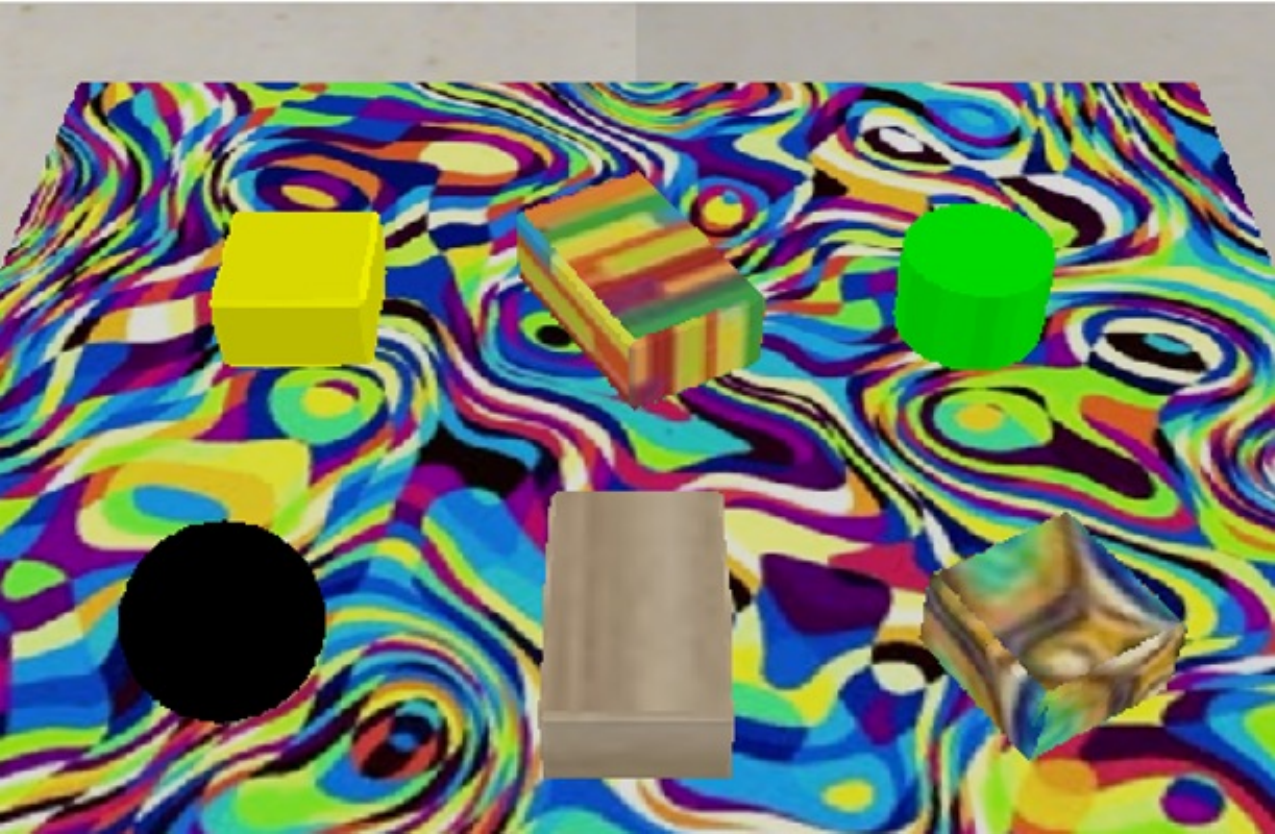}
  \end{subfigure}
  \caption{\textbf{Examples of basic objects}. Synthetic objects of various colors and shapes are used for learning object attributes and grasping affordances. To ensure shape attribute learning, we include objects having random textures.}
  \label{fig:basic}
\end{figure}
\begin{algorithm}[!t]
\caption{Online Data Collection}
\textbf{Initialize} bounded buffer $B$\\
\textbf{Notations}: $\epsilon$-greedy policy $\pi_\epsilon$, our model $\phi$, image $v$, text $t$, mask $M$, action $a$, and label $\bar{q}_e$
\begin{algorithmic}[1]
\While{collecting data}
\State reset the simulation and randomly drop basic objects
\State get image $v_{\text{pre}}$ and randomly choose a command $t$
\State execute action $a \gets \pi_{\epsilon}(\phi; v_{\text{pre}}, t, \epsilon)$
\State label $\bar{q}_e$ according to grasping results
\State save $v_{\text{pre}}$, $t$, $M$, $a$ and $\bar{q}_e$ into $B$
\If{successful grasp}
\State save $v_{\text{post}}$ into $B$ with HER
\EndIf
\State sample a batch from $B$ to train the model
\EndWhile
\end{algorithmic}
\label{alg:training}
\end{algorithm}
To achieve self-supervision, we create a simulation environment in which objects are identified and grasped based on a description of semantic attributes. We collect training data in simulation with the following procedure, as summarized in Algorithm \ref{alg:training}. Several objects are randomly dropped into the workspace in front of the robot. Given a workspace image and a query text, the robot learns to grasp a target under $\epsilon$-greedy exploration \cite{sutton2018reinforcement} ($\epsilon\!=\!0$ during testing, i.e., an argmax policy). We save the workspace images, query text, background masks, executed actions, and results into a bounded buffer. The ground-truth labels are automatically generated for learning grasping affordances. The label $\bar{q}_e$ in (\ref{eq:motion}) is assigned as the attribute similarity in (\ref{eq:similarity}) between the query text and the grasped object (0 if no object grasped). We also save the workspace image after a successful grasping for learning object attributes. To deal with sparse rewards in target-driven grasping, the hindsight experience replay (HER) technique \cite{andrychowicz2017hindsight} is applied.  If a non-target is grasped, we add the additional positive sample by relabeling the target text. Fig. \ref{fig:basic} shows the basic objects of various colors (red, green, blue, yellow, and black) and shapes (cube, cuboid, cylinder, and sphere) used in our simulation. We choose these colors and shapes because they are foundational for common objects in daily life. To enrich the distribution of training data, we perturb color RGB values, randomize sizes and heights of the objects, and randomize textures of the workspace. Using the domain randomization techniques \cite{tobin2017domain}, we can generate a number of randomized properties in simulation and achieve a model with better generalization. At every iteration, we sample a batch from the buffer and run one off-policy training. The training loss is defined as
\begin{align}
\label{eq:train_loss}
    \mathcal{L}_{train} = \mathcal{L}_{grasp} + \lambda_{a} \mathcal{L}_{attr}
\end{align}
by combining both motion loss $\mathcal{L}_{grasp}$ (\ref{eq:motion}) and metric loss $\mathcal{L}_{attr}$ (\ref{eq:metric}). We train the proposed model using stochastic gradient descent with a learning rate of $10^{-4}$, momentumn of $0.9$, and weight decay of $2\times10^{-5}$ for 5k iterations. Each training iteration involves capturing data, computing a forward pass, executing an action, and backpropagating. After collecting a dataset of 5k samples, we replay the entire data for 100 epochs.
\begin{figure}[!t]
  \centering
  \begin{subfigure}{0.23\textwidth}
    \includegraphics[width=\textwidth]{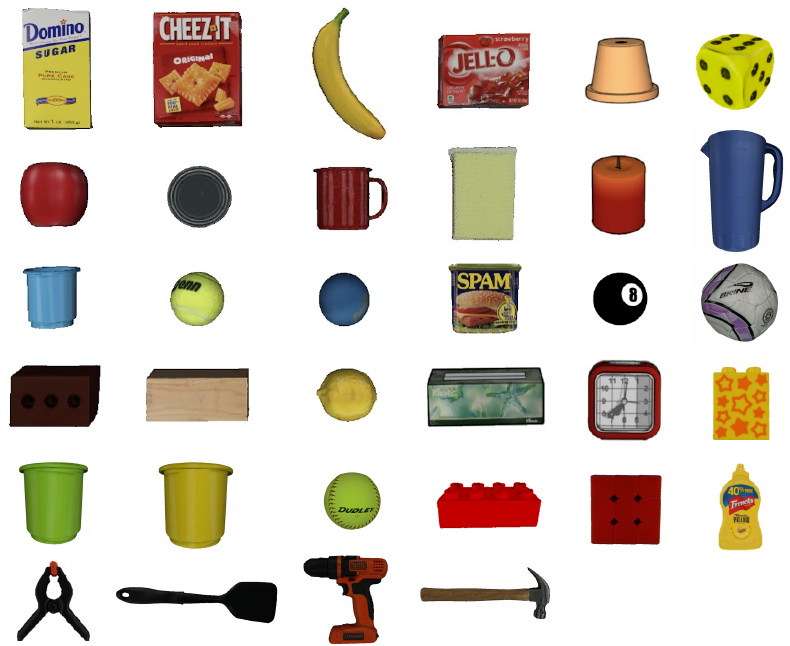}
    \caption{Simulated novel objects}
    \label{fig:testing_sim}
  \end{subfigure}
  \hfill
  \begin{subfigure}{0.25\textwidth}
    \includegraphics[width=\textwidth]{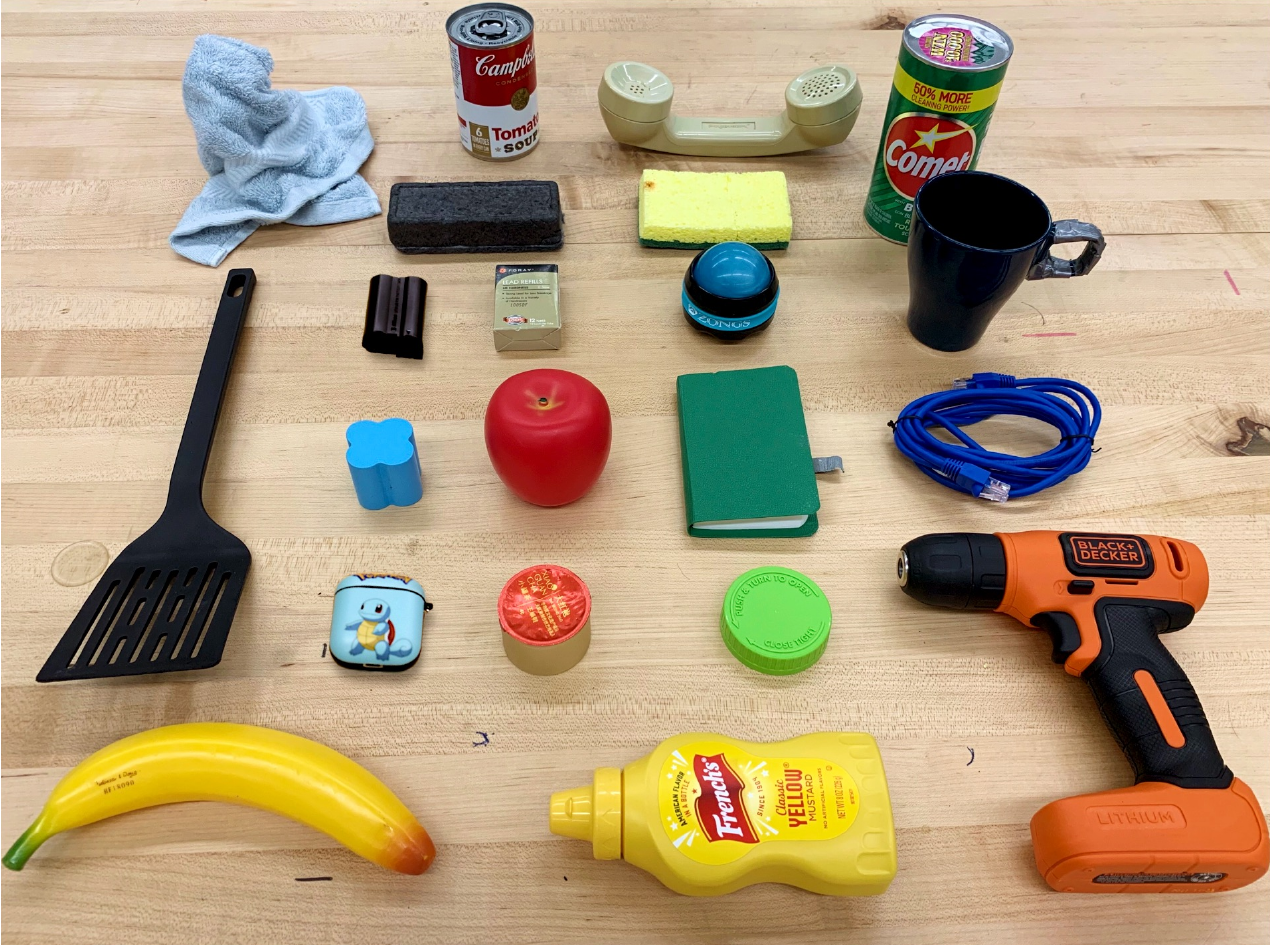}
    \caption{Real-world objects}
    \label{fig:testing_real}
  \end{subfigure}
  \caption{\textbf{Testing objects} in simulation and the real world. We use the testing objects that share similar attributes with the training objects. See Appendix for more details.} 
  \label{fig:testing}
\end{figure}

\section{Data-Efficient Adaptation}
\begin{figure}[!t]
  \centering
  \begin{subfigure}{0.48\textwidth}
    \includegraphics[width=\textwidth]{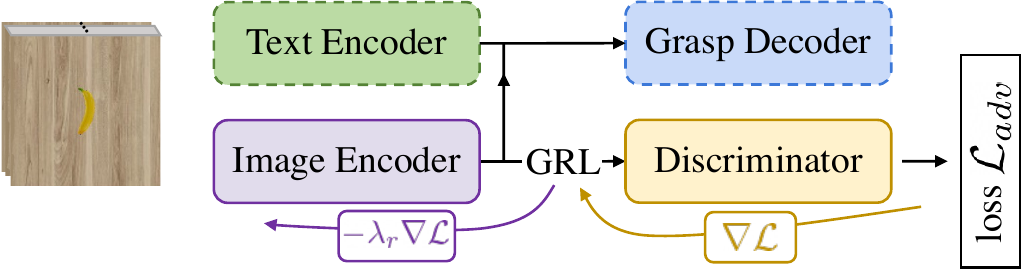}
    \vspace{-18pt}
    \caption{Adversarial Adaptation}
    \label{fig:adapt_pipe_adv}
  \end{subfigure}
  \vfill
  \vspace{10pt}
  \begin{subfigure}{0.48\textwidth}
    \includegraphics[width=\textwidth]{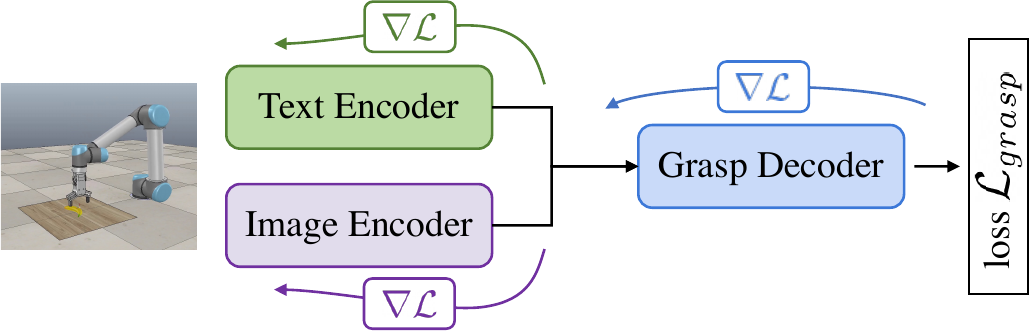}
    \caption{One-Grasp Adaptation}
    \label{fig:adapt_pipe_1grasp}
  \end{subfigure}
  \caption{\textbf{Overview of the proposed adaptation schemes}. In adversarial adaptation (top), the adversarial loss $\mathcal{L}_{adv}$ regulates the image encoder using augmented data of unlabeled images to learn domain-invariant features for the decoder. The dotted boxes indicate no gradient flows corresponding to $\mathcal{L}_{adv}$ (there are gradients from $\mathcal{L}_{train}$ though). One-grasp adaptation updates the end-to-end model through $\mathcal{L}_{grasp}$ using augmented data from one grasp trial.}
  \label{fig:adapt_pipe}
\end{figure}
Due to the high cost of collecting data on real robots, we often choose to train robotic models in a simulator. However, the domain gap between the source domain (e.g., simulation, trained objects) and the target domain (e.g., the real world, novel objects) frequently leads to the failure of the learned models. We propose to, in addition to randomizing the source domain in Sec. \ref{subsec:method_data}, adapt our learned model using data from the target domain to further alleviate the domain shifts. One typical adaptation approach is fine-tuning the pre-trained model. However, the fine-tuning methods remain expensive in terms of data usage. In this section, as shown in Fig. \ref{fig:adapt_pipe}, we propose two data-efficient adaptation methods: 1) adversarial adaptation, which adapts the image encoder using unlabeled images, and 2) one-grasp adaptation, which updates the end-to-end model using one grasp trial. The two adaptation methods can be either used independently or in combination for performance improvement.

\subsection{Adversarial Adaptation}\label{subsec:method_adv}
Despite that our generic model trained using the simulated basic objects shows good generalization (see Sec. \ref{subsec:exp_grasp}), the visual feature shifts (e.g., objects, lighting conditions, and scene configurations, etc.) are inevitable. As a result, the image encoder is likely to produce out-of-distribution visual embeddings, leading to the failure of the grasping model. To reduce the influence of the domain shifts, we propose to use adversarial adaptation \cite{tzeng2017adversarial} to learn domain-invariant visual features that are transferable across different domains. In our problem setup, the simulated basic objects constitute the source domain, and our goal is to transfer the learned model to a target domain that is prone to domain shifts.

As shown in Fig. \ref{fig:adapt_pipe_adv}, adversarial adaptation regularizes the weights of the image encoder $\phi_v$ by enforcing a two-player game similar to the generative adversarial network (GAN) \cite{goodfellow2014generative}. A domain classifier (i.e., discriminator) learns to distinguish between two domains, while the image encoder learns to fool the domain classifier by learning domain-invariant features. To achieve adversarial training, we connect the encoder and the discriminator via a gradient reversal layer (GRL) \cite{ganin2016domain} that has reverse forward and back-propagation schemes. The GRL $\mathcal{R}$ is an identity mapping during forward-propagation but reverses the sign of the gradients during back-propagation:
\begin{align}
    &\mathcal{R}(x) = x \\
    &\frac{d \mathcal{R}}{d x} = -\lambda_rI
\end{align}
where $I$ is an identity matrix, and $\lambda_r$ is a positive constant.

\begin{figure}[!t]
  \begin{subfigure}[t]{0.152\textwidth}
    \includegraphics[width=\textwidth]{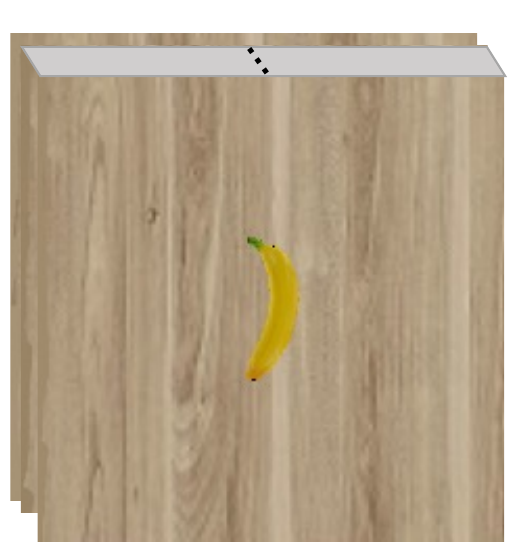}
    \caption*{object images}
  \end{subfigure}
  \begin{subfigure}[t]{0.152\textwidth}
    \includegraphics[width=\textwidth]{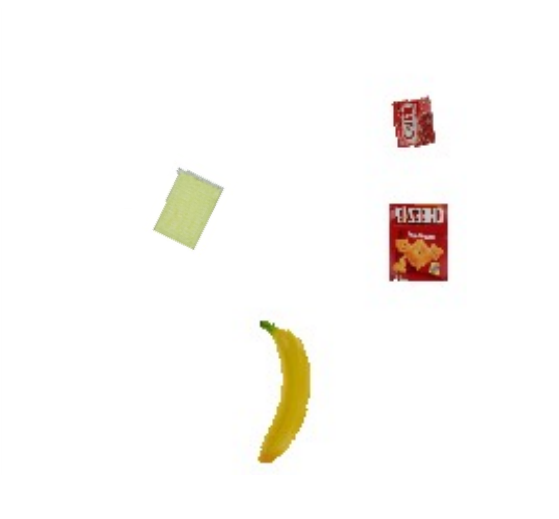}
    \caption*{extracted objects}
  \end{subfigure}
  \begin{subfigure}[t]{0.152\textwidth}
    \includegraphics[width=\textwidth]{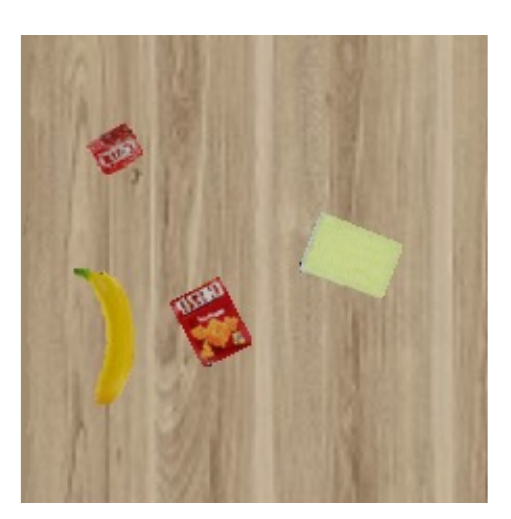}
    \caption*{augmented image}
  \end{subfigure}
  \caption{\textbf{Object-level data augmentation}. We extract objects from the images of a single object, apply the augmentation methods (e.g., shifting, rotation), and assemble them on a background to generate diverse adaptation data for adversarial adaptation.}
  \label{fig:obj_aug}
\end{figure}

During adaptation, images from the two different domains are fed into the image encoder. The domain classifier $f_d$ takes as input the encoded features, and is trained to predict which domain the feature is from, by maximizing the binary cross-entropy $\mathcal{H}_d$. Using the source domain dataset $\mathcal{D}_\mathpzc{s}=\{v_\mathpzc{s}, t_\mathpzc{s}, y_\mathpzc{s}\}^{N_\mathpzc{s}}$ ($\{$image, text, label$\}$ collected in Algorithm \ref{alg:training}) and assuming a target domain dataset $\mathcal{D}_\mathpzc{t}=\{v_\mathpzc{t}\}^{N_\mathpzc{t}}$, the model is updated to be optimal on the training loss $\mathcal{L}_{train}$ (\ref{eq:train_loss}) under the regularization of the adversarial loss $\mathcal{L}_{adv}$
\begin{align} \label{eq:adva}
    \mathcal{L}_{adv} &= -\sum_{v_i \in \mathcal{D}_\mathpzc{s} \cup \mathcal{D}_\mathpzc{t}} \mathcal{H}_{d}(f_d(\mathcal{R}(\phi_{v, \text{vec}}(v_i))), d_i)\\
    \mathcal{L}_{advadp} &= \mathcal{L}_{train}(v_\mathpzc{s}, t_\mathpzc{s}, y_\mathpzc{s}) + \mathcal{L}_{adv}(v_\mathpzc{t})
\end{align}
where $\mathcal{L}_{advadp}$ is the overall loss for adversarial adaptation, $d_i \in \{0, 1\}$ are the binary domain label for each input $v_i$. Note that the reversal layer $\mathcal{R}$ is augmented between the classifier $f_d$ and the encoder $\phi_{v, \text{vec}}$ and updates them in reversal directions.

The target dataset $\mathcal{D}_\mathpzc{t}$ for the new domain is the prerequisite to performing adversarial adaptation. As shown in Fig. \ref{fig:obj_aug}, we propose a object-level augmentation (\textit{ObjectAug}) approach instead of collecting data from a vast number of configurations. Since the grasping label is not required in $\mathcal{D}_\mathpzc{t}$, synthetic generation of a large image dataset would be more efficient. We begin by collecting RGB-D images of all conceivable objects in the target domain. Using the object mask acquired from the depth channel, we extract each object individually and randomize them with the single-object augmentation methods commonly used (e.g., scaling, flipping, and rotation). To generate an augmented RGB-D image, the augmented objects are randomly sampled and shifted before being overlaid with a background image. We also perform IoU threshold verification to avoid dense overlapping. To simulate varying conditions of target domains, we can apply the visual jitter technique discussed in Sec. \ref{subsec:method_syssim} to obtain more diverse data. Given the ease with which unlabeled images may be acquired (e.g., the internet, image collection), generating a target dataset that is of the same magnitude as the source dataset is rather efficient.

\subsection{One-Grasp Adaptation}\label{subsec:method_1grasp}
By learning domain-invariant features, the adversarial adaptation technique in Sec. \ref{subsec:method_adv} improves model generalization using unlabeled images of the target domain. However, the adversarial loss uses unlabeled images to only update the image encoder and leave the text encoder and the grasping affordance decoder unadapted. When deploying in a new domain, end-to-end model fine-tuning is often necessary, but this comes at the cost of a large dataset covering all potential testing object configurations. To further adapt to novel objects and new scenes in a data-efficient manner, we present a one-grasp adaptation scheme (see Fig. \ref{fig:adapt_pipe_1grasp}) that only requires one successful grasp of a novel object. The inductive bias of object attributes in Sec. \ref{subsec:method_metric} is the key to adaptation in this limited-data regime. If similar objects are enforced closer in the embedding space, the adaptation distance for a novel object is likely to be shorter \cite{laradji2020m}.

\begin{figure}[!t]
  \begin{subfigure}[t]{0.17\textwidth}
    \includegraphics[width=\textwidth]{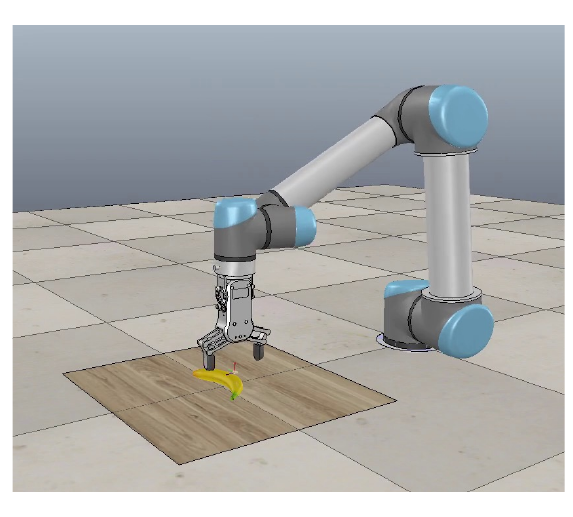}
    \caption*{one-grasp collection}
  \end{subfigure}
  \begin{subfigure}[t]{0.152\textwidth}
    \includegraphics[width=\textwidth]{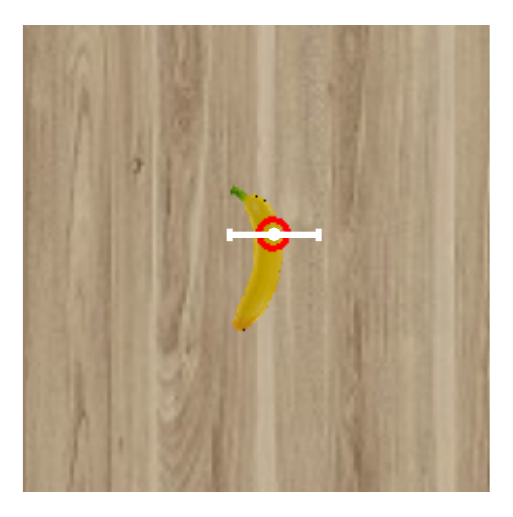}
    \caption*{collected data}
  \end{subfigure}
  \begin{subfigure}[t]{0.152\textwidth}
    \includegraphics[width=\textwidth]{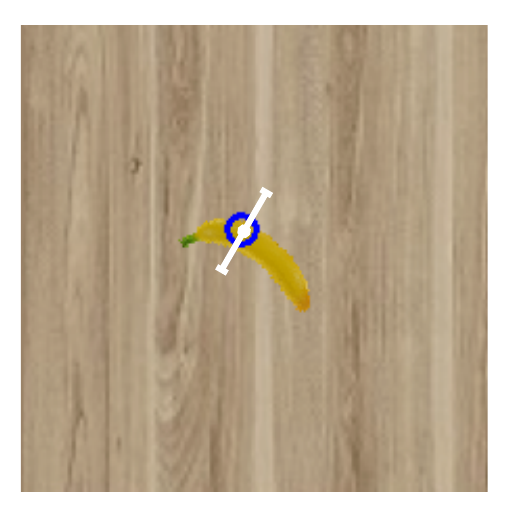}
    \caption*{augmented data}
  \end{subfigure}
  \caption{\textbf{One-grasp data augmentation}. We place the target object solely on the workspace and collect one successful grasp. To enrich the distribution of grasping angles, we rotate the collected data and synthesize a few samples for one-grasp adaptation. One rotation example of the augmented data is visualized.}
  \label{fig:adapt}
\end{figure}

The proposed one-grasp adaptation method improves the model performance on a novel target object at the cost of only one grasp. The adaptation data is collected with the following one-grasp data augmentation (\textit{OneGraspAug}) procedure. We place the object solely in the workspace and run the generic model to collect one successful grasp. The setting of a sole object facilitates grasping and avoids combinatorial object arrangements. Because convolutional neural networks are not rotation-invariant by design, we also augment the grasp data by rotating with various orientations to achieve rotation-invariance \cite{jaderberg2015spatial}, \cite{quiroga2018revisiting}, i.e., the ability to recognize and grasp an object regardless of its orientation. As shown in Fig. \ref{fig:adapt}, we rotate the collected image and action execution to have rotated versions of the collected data.

In the adaptation stage, we add the category name of the object as an additional token to the query text, e.g., ``apple, red sphere'' for the testing object apple. The token embedding of the object name is initialized properly to keep the embedding vector of the query text unchanged. The addition of the object name allows for a more specific grasping instruction and distinguishing from similar objects via adaptation. By optimizing over motion loss $\mathcal{L}_{grasp}$ in (\ref{eq:motion}), we jointly fine-tune the recognition and grasping of our model for unknown objects and scenes. As delineated in Sec. \ref{subsec:exp_adapt}, the adapted model outputs higher affordances on the target objects that are not seen and grasped before.

\section{System Implementation}
\subsection{Simulation Environment}\label{subsec:method_syssim}
We use CoppeliaSim \cite{rohmer2013v} to build our simulation environment. The simulation setup includes a UR5 robot arm and an RG2 gripper, with Bullet Physics 2.83 for dynamics and CoppeliaSim’s internal inverse kinematics module for robot motion planning. We simulate a statically mounted overhead 3D camera in the environment from which perception data is captured. The camera renders RGB-D images with a $640 \times 480$ pixels resolution using OpenGL. We use various basic and novel objects for training and testing in the simulation. As shown in Fig. \ref{fig:basic}, the basic objects consist of 36 different 3D toy blocks, whose shapes and colors are randomly chosen during experiments. We collect 34 different 3D household objects from the YCB dataset \cite{calli2015ycb} or the 3D mesh library SketchUp \cite{chopra2012introduction}, as shown in Fig. \ref{fig:testing_sim}. To produce more diverse simulation configurations, the simulation environment supports several domain randomization techniques, including background randomization, color jitter (i.e., randomly changing the brightness, contrast, and saturation of the color channel), and depth jitter (i.e., adding Gaussian noise to the depth channel).

\subsection{Real-Robot System}
Fig. \ref{fig:intro_a} shows our real-world setup that includes a Franka Emika Panda robot arm with a FESTO DHAS soft gripper and a hand-mounted Intel RealSense D415 camera overlooking a tabletop scene. We use the soft fingers because they are more suitable for grasping the objects in our experiments and are similarly compliant to the RG2 fingers. For perception data, RGB-D images of resolution $640 \times 480$ are captured from the RGB-D camera, statically mounted on the robot arm. The camera is localized with respect to the robot base using the automatic calibration procedure of ViSP \cite{marchand2005visp}, during which the camera tracks the location of a checkerboard pattern taped on the table. For a given pose, the robot follows the corresponding trajectory generated with MoveIt \cite{chitta2012moveit} in open-loop. The entire system is implemented under the Robot Operating System (ROS) framework and runs on a PC workstation with an Intel i7-8700 CPU and an NVIDIA 1080Ti GPU. Objects vary throughout tests, with a collection of 20+ different household objects being used to test model generalization to novel objects, as shown in Fig. \ref{fig:testing_real}. 

\section{Experiments}
We propose training with simulated basic objects first to have a generic model and then adapting it to novel objects and real-world scenes. In the experiments, we first analyze the structured metric space of our generic model and show the consistency between attention and grasping maps. Next, we evaluate the instance grasping performance of the generic model and show its modest generalization even before adaptation. Then, we adapt the model using the proposed adversarial and one-grasp adaptation methods and test the grasping models after adaptation. Finally, we run a series of ablation studies to investigate the two adaptation methods. The goals of the experiments are four-fold:
\begin{itemize}[noitemsep, topsep=0pt]
    \item[1)] to show the effectiveness of multimodal attribute learning for instance robotic grasping,
    \item[2)] to evaluate our attribute-based grasping system in both simulated and real-world settings,
    \item[3)] to evaluate the proposed adversarial and one-grasp adaptation methods, and 
    \item[4)] to show the importance of the proposed data augmentation methods for grasping adaptation.
\end{itemize}

\subsection{Multimodal Attention Analysis}\label{subsec:exp_metric}
\begin{figure}[!t]
    \centering
    \begin{subfigure}{0.48\textwidth}
        {\includegraphics[width=0.24\textwidth]{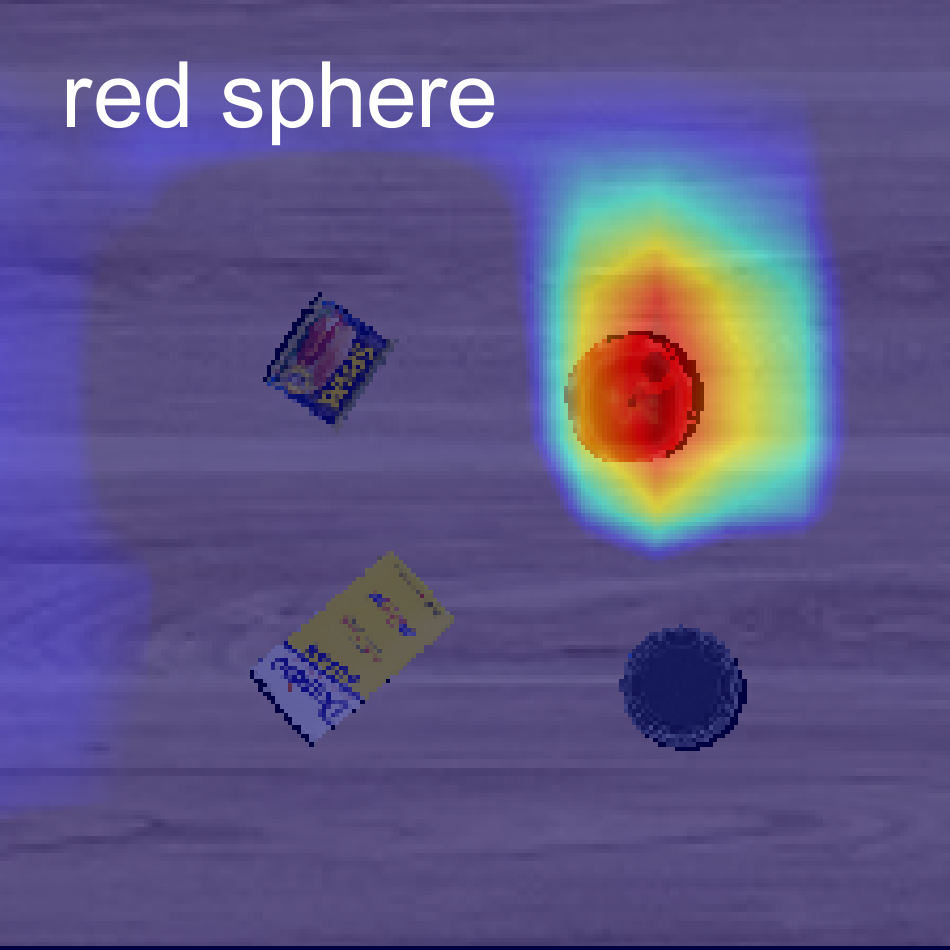}}\hfill
        {\includegraphics[width=0.24\textwidth]{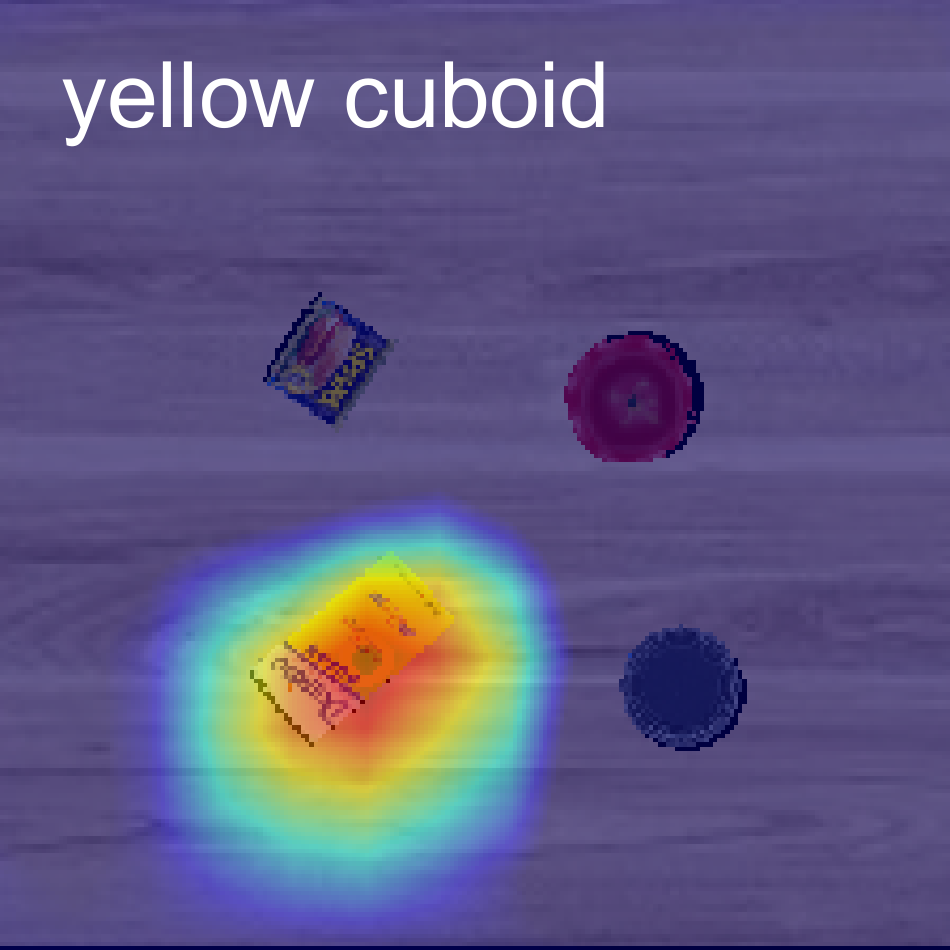}}\hfill
        {\includegraphics[width=0.24\textwidth]{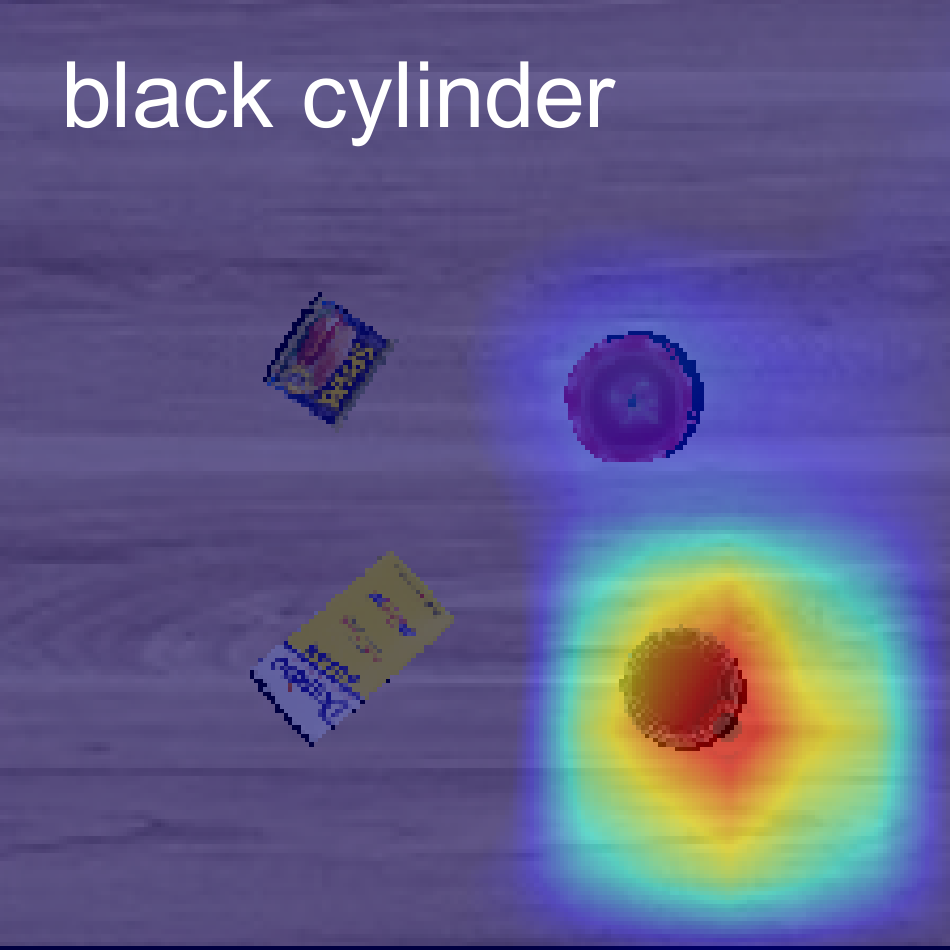}}\hfill
        {\includegraphics[width=0.24\textwidth]{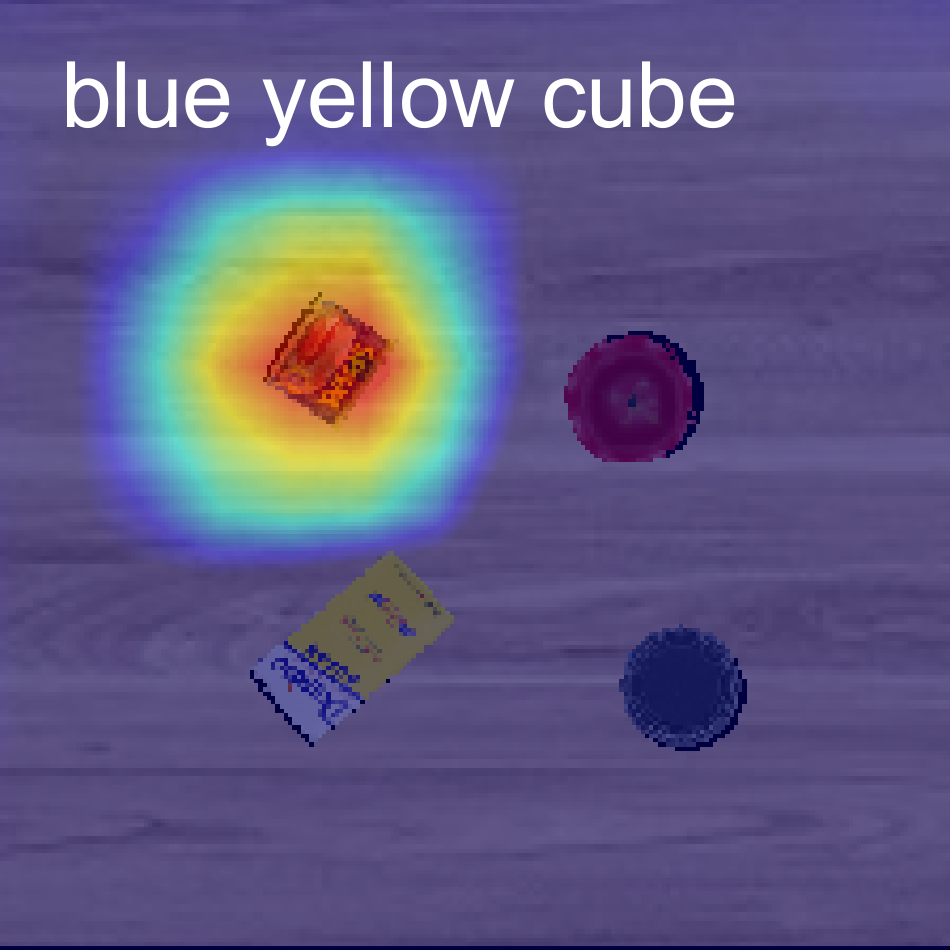}}%
        \caption{Encoder's attention}
        \vspace{2pt}
        \label{fig:metric}
    \end{subfigure}
    \begin{subfigure}{0.48\textwidth}
        {\includegraphics[width=0.24\textwidth]{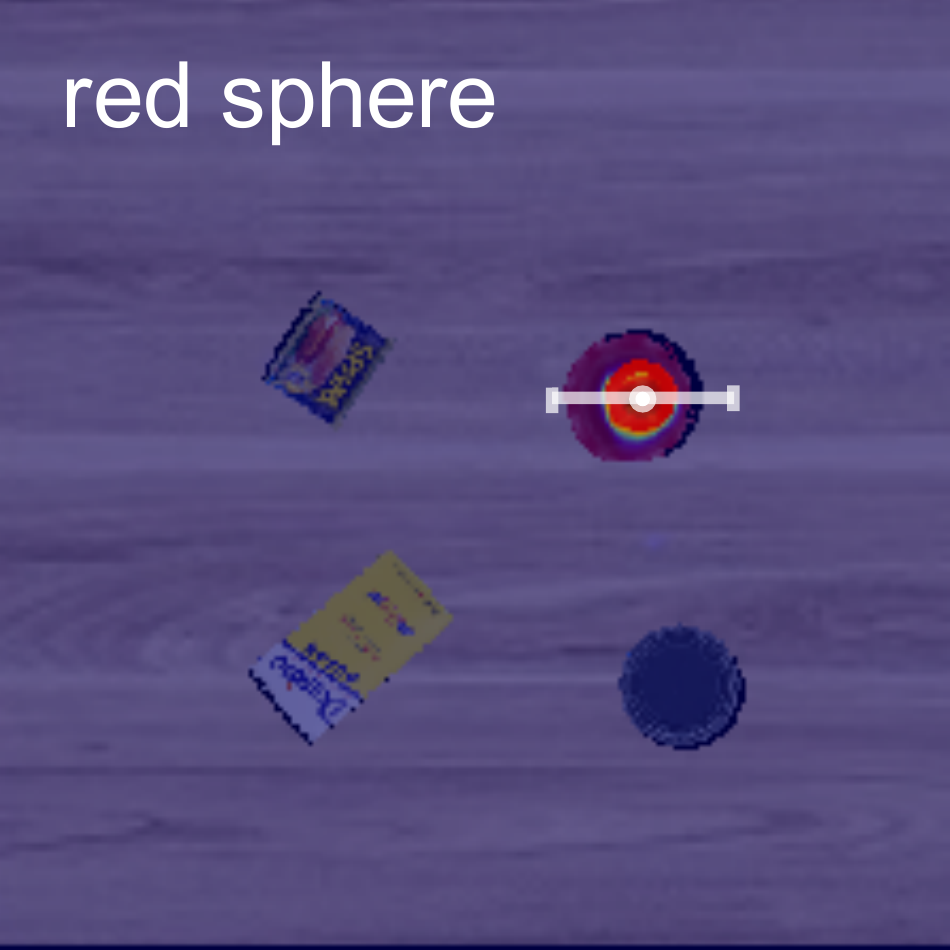}}\hfill
        {\includegraphics[width=0.24\textwidth]{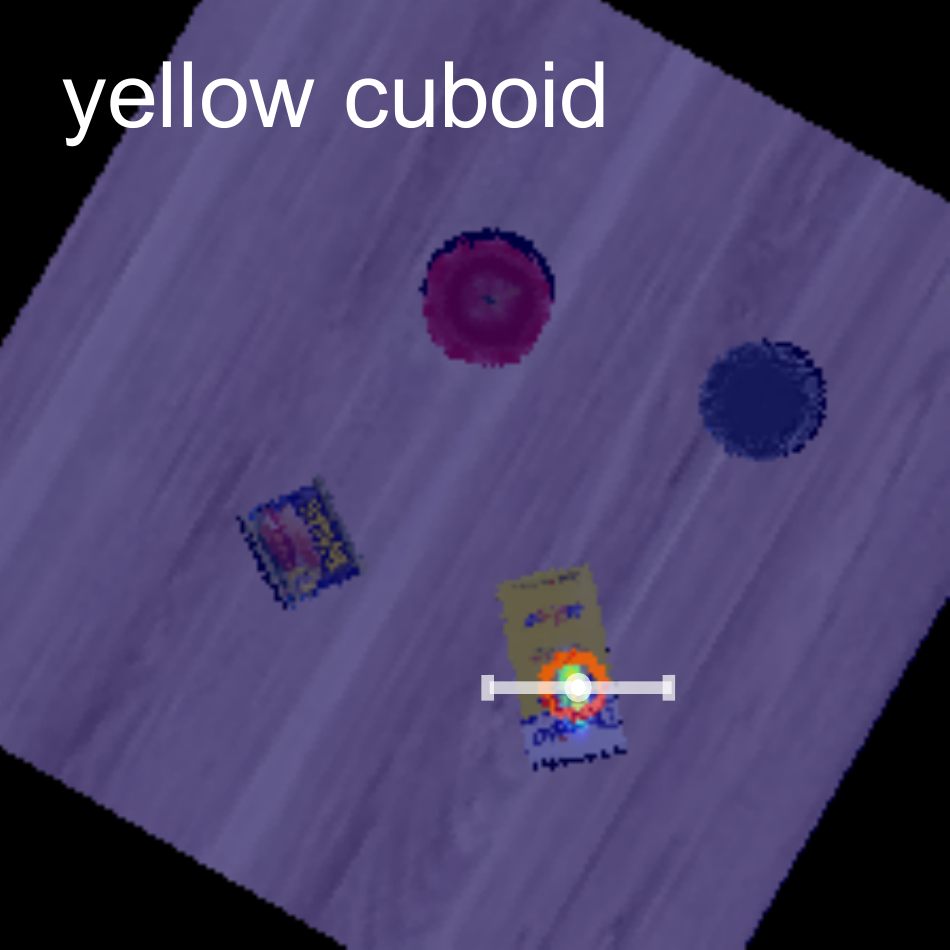}}\hfill
        {\includegraphics[width=0.24\textwidth]{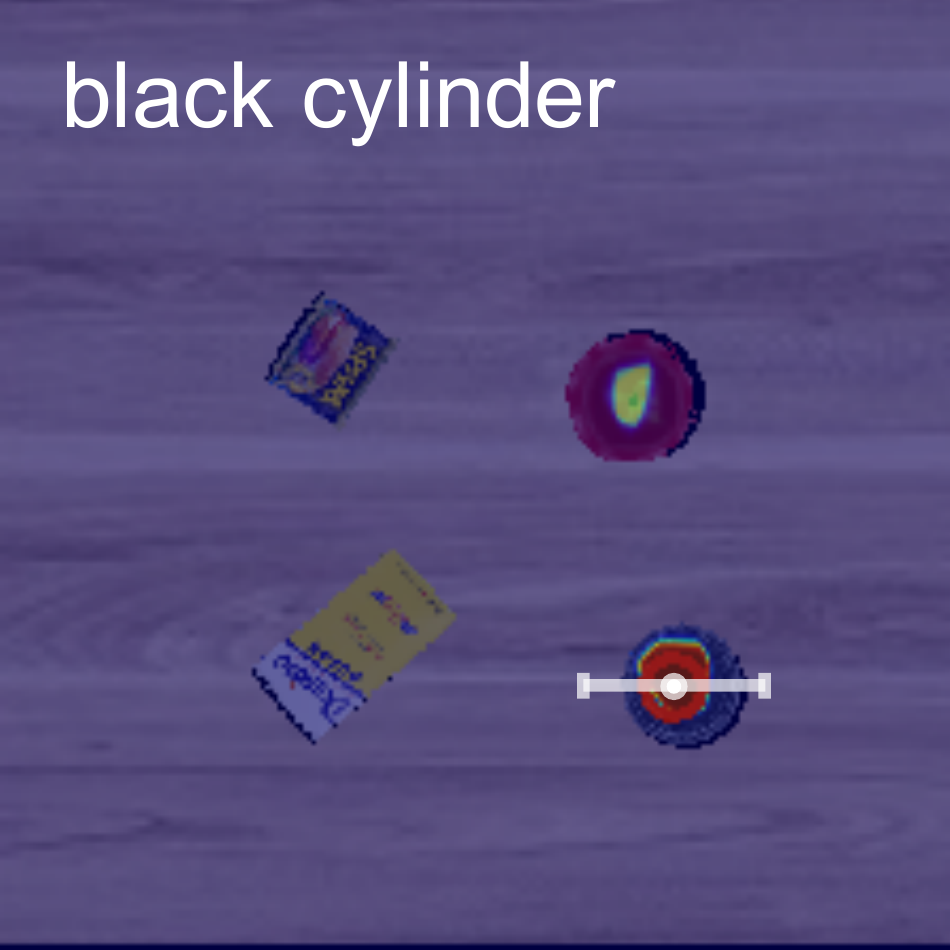}}\hfill
        {\includegraphics[width=0.24\textwidth]{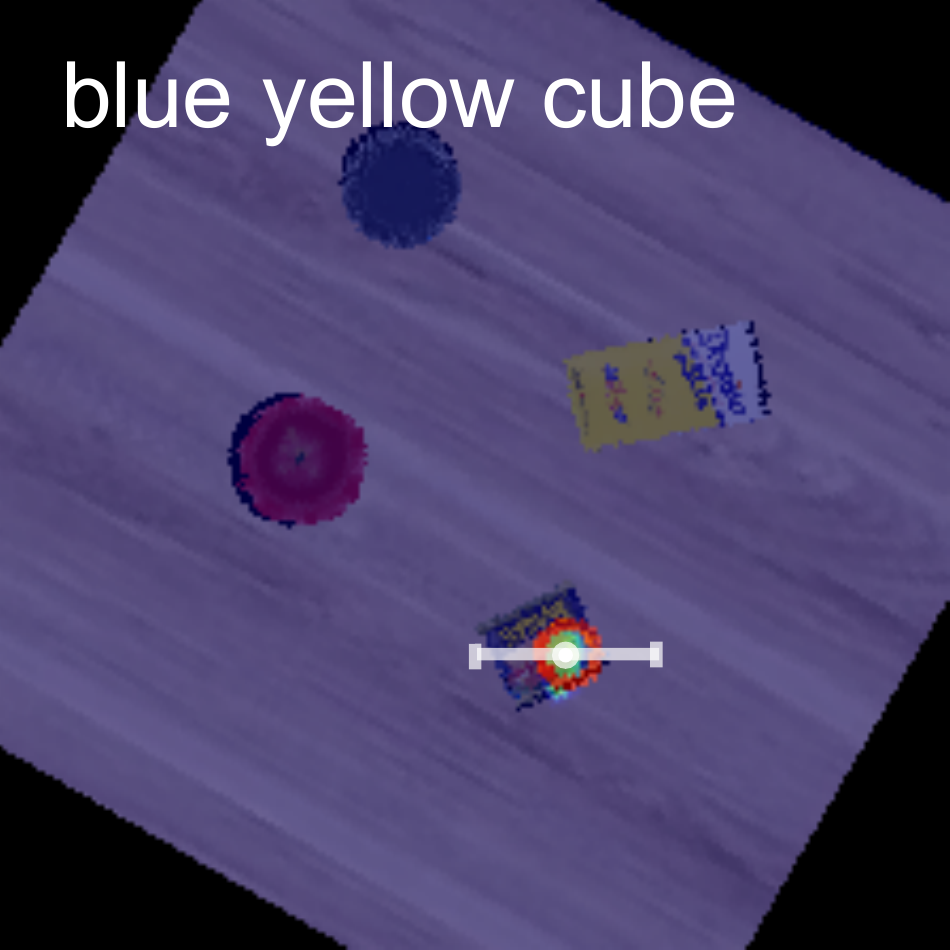}}%
        \caption{Grasping affordances}
        \label{fig:grasp}
    \end{subfigure}
    \caption{\textbf{Visualization of attention and grasping.} (\subref{fig:metric}) shows attention of our encoder, and (\subref{fig:grasp}) shows heatmaps of grasping affordances for different targets (described by the query text). The maps of attention and grasping are consistent even for the novel objects.}%
    \label{fig:sim}
\end{figure}
By embedding workspace images and query text into a joint metric space, the multimodal encoder ($\phi_v$ and $\phi_t$), supervised by metric loss $\mathcal{L}_{attr}$ and motion loss $\mathcal{L}_{grasp}$, learns attending to text-correlated visual features. We visualize what our model ``sees'' by computing the dot product of text vector $\varphi_t$ with each pixel of the visual matrix $\varphi_{v, spa}$. This computation obtains an attention heatmap over the image, which refers to the similarity between the query text and each pixel’s receptive field (see Fig. \ref{fig:metric}). We quantitate the attention of our model and report its attention localization performance in Table \ref{tab:acc} (see {\textit{Ours-Attention}}). \textbf{Evaluation metrics:} An attention localization is considered correct only if the maximum value in the attention heatmap lies on the target object.

\textit{Ours-Attention} (in Table \ref{tab:acc}) performs target localization at a 74.5\% accuracy on simulated novel objects and a 70.2\% accuracy on real-world objects, without any localization supervision provided. In summary, our multimodal embeddings demonstrate a consistent pattern across object categories and scenes. Though the localization results are not directly used for grasping, the consistent embeddings facilitate learning, generalization, and adaptation of our grasping model, as shown in Fig. \ref{fig:sim} and discussed in the following subsections.

\subsection{Generic Instance Grasping}\label{subsec:exp_grasp}
We compare the instance grasping performance of our generic model with the following baselines:
\begin{itemize}[noitemsep, topsep=0pt]
    \item[1)] \textit{Indiscriminate} is an indiscriminate grasping version of our approach and composed of a visual spatial encoder $\phi_{v, \text{spa}}$ and a grasping affordances decoder $\phi_g$. We collect a dataset of binary indiscriminate grasping labels and train \textit{Indiscriminate} using $\mathcal{L}_{grasp}$ in (\ref{eq:motion}).
    \item[2)] \textit{ClassIndis} extends \textit{Indiscriminate} with an attributes classifier that is trained to predict color and shape attributes on cropped object images. We filter the grasping maps from \textit{Indiscriminate} using the mask of a target recognized by the classifier.
    \item[3)] \textit{EncoderIndis} is similar to \cite{cohen2019grounding} and is another extension of \textit{Indiscriminate}, which leverages a multimodal encoder ($\phi_{v, \text{vec}}$ and $\phi_t$ in Sec. \ref{subsec:method_grasp}) for text template matching. The encoder is trained using $\mathcal{L}_{attr}$ in (\ref{eq:metric}) to evaluate the similarity between each cropped object image and query text. During training, we also include attributes classification as an axillary task. 
    \item[4)] \textit{AttrID} is for an ablation study of our text encoder $\phi_t$. The only difference between \textit{AttrID} and the proposed method is that \textit{AttrID} takes the attribute shape and color ID one-hot encoding as the  system input, but our generic model uses Word2Vec continuous bag-of-words (CBOW) model to convert the texts into vector inputs. During training, we use both the motion loss and the metric loss to update the model.
    \item[5)] \textit{NoMetric} is for an ablation study of multimodal metric loss. We simply remove the metric loss on the basis of our approach during its training.
\end{itemize}
\textbf{Evaluation metrics:} These methods have different target recognition schemes: \textit{ClassIndis} and \textit{EncoderIndis} recognize a target by classification and text template matching respectively; \textit{NoMetric}, \textit{AttrID} and \textit{Ours} are end-to-end. We report their target recognition performance (in addition to instance grasping performance, as in the next paragraph). A target localization is correct only if the predicted grasping location lies on the target object. The instance grasping success rate is defined as $\frac{\text{\# of successful grasps on correct target}}{\text{\# of total grasps}}$. In each testing scene, we only execute grasping once.
%
%

\begin{table}[!t]
    \centering
    \caption{Target Recognition Accuracy (\%)}
    \label{tab:acc}
    \begin{tabular}{c|c|c|c}
    \hline
    Method & sim basic & sim novel & real novel\\
    \hline
    ClassIndis & 100.0 & {56.3} & {35.7}\\
    EncoderIndis & 95.8 & {71.5} & {57.1} \\
    NoMetric & 93.4 & {70.4} & {54.0}\\
    Ours-Attention & 95.3 & {74.5} & {70.2}\\
    {AttrID} & {96.8} & {79.1} & {70.6}\\
    Generic (Ours) & \textbf{100.0} & {\textbf{79.7}} & {\textbf{71.8}}\\
    \hline
    \end{tabular}
    \vspace{13pt}
    \centering
    \caption{Instance Grasping Success Rate (\%)}
    \label{tab:grasp}
    \begin{tabular}{c|c|c|c}
    \hline
    Method & sim basic & sim novel & real novel\\
    \hline
    Indiscriminate & 27.2 & 22.8 & 13.5\\
    ClassIndis & 91.6 & 51.0 & 32.9\\
    EncoderIndis & 89.1 & 63.9 & 52.0 \\
    NoMetric & 90.5 & 62.7 & 50.0\\
    AttrID & 90.8 & 68.7 & 60.0\\
    Generic (Ours) & \textbf{98.4} & \textbf{72.1} & \textbf{63.1}\\
    \hline
    \end{tabular}
\end{table}

We evaluate the methods on both simulated basic (sim basic) and simulated novel (sim novel) objects in simulation, where there are 1200 test cases for the basic objects (Fig. \ref{fig:basic}) and 3400 test cases for the 34 novel objects (Fig. \ref{fig:testing_sim}, mostly from the YCB dataset \cite{calli2015ycb}). We assume the objects are placed right-side up to be stable while their 4D pose (3D position and a yaw angle) can vary arbitrarily. For each testing object, we pre-choose a query text that best describes its color and/or shape. In each test case, four objects are randomly sampled and placed in the workspace, except avoiding any two objects with the same attributes. The robot is required to grasp the target queried by an attribute text. We report the results of target recognition in Table \ref{tab:acc} and the results of instance grasping in Table \ref{tab:grasp}.

Overall, our approach outperforms the baselines remarkably (in both recognition and grasping) and achieves a 98.4\% grasping success rate on the simulated basic objects and an 72.1\% grasping success rate on the simulated novel objects. \textit{ClassIndis} extends \textit{Indiscriminate} that is well trained in target-agnostic tasks and performs well on the basic objects, but the attributes classifier generalizes poorly. \textit{EncoderIndis} utilizes a more generalizable recognition module and performs better on the novel objects. However, \textit{EncoderIndis} fails to reach optimality because its separately-trained recognition and grasping modules have different training objectives from instance grasping. Despite training the recognition and grasping modules simultaneously, \textit{AttrID} using sparse attribute one-hot IDs as a substitute for text inputs yields lower recognition accuracy and grasping success rate compared to \textit{Ours}. As an ablation study, the performance gap between \textit{NoMetric} and \textit{Ours} shows the effectiveness of multimodal metric loss, which supervises the joint latent space to produce consistent embeddings, as discussed in Sec. \ref{subsec:exp_metric}. Our approach successfully learns object attributes that generalize well to novel objects, as shown in Fig. \ref{fig:sim}.

We further evaluate our approach and the baselines on the real robot before any adaptation (see Table \ref{tab:acc} and \ref{tab:grasp}). Fig. \ref{fig:testing_real} shows 21 testing objects of various colors and shapes used in our real-robot experiments. The robot is tasked to grasp the target within a combination of 6 objects placed on the table. We use the same 21 object combinations that are randomly generated and repeat each combination twice, resulting in a total of 252 grasping trials for each method. Overall, the grasping performance of all the methods decreases due to the domain gap. However, our approach shows the best generalization and achieves a 63.1\% grasping success rate, before adaptation, in the real-world scenes.
%
%

\subsection{Adapted Instance Grasping}\label{subsec:exp_adapt}

The generic model in Sec. \ref{subsec:exp_grasp} infers the object closest to the query text as the target. Overall, our generic model demonstrates good generalization despite the gaps in the testing scenes. Specifically, these gaps are 1) RGB values of the testing objects deviate from training ranges, 2) some testing objects are multi-colored, 3) shape and size differences between the testing objects and the training objects, and 4) depth noises in the real world causing imperfect object shapes. To account for the gaps, we further adapt our generic model to increase instance recognition and grasping performance.

We first collect one successful grasp of a solely placed target object and then augment the collected data by rotating with additional $N-1$ angles, as discussed in Sec. \ref{subsec:method_1grasp} and shown in Fig. \ref{fig:adapt}. The compared methods that are adapted with the same adaptation data are as follows:
\begin{itemize}[noitemsep, topsep=0pt]
    \item[1)] \textit{ClassIndis} updates its attributes classifier for a better recognition accuracy on the adaptation data. 
    \item[2)] \textit{EncoderIndis} minimizes the latent distance between cropped target images and query text to improve text template matching.
    \item[3)] \textit{NoMetric} takes as input images and text, and minimizes motion loss on the adaptation data.
    \item[4)] \textit{One-Grasp} is our prior work \cite{yang2021attribute} which updates the encoder-decoder in an end-to-end manner.
    \item[5)] \textit{AttrID-One-Grasp} uses the same adaptation method and data as \textit{One-Grasp} baseline on \textit{AttrID} from Sec. \ref{subsec:exp_grasp}.
\end{itemize}

\begin{table}[!t]
    \centering
    \caption{Adapted Instance Grasping (\%)}
    \label{tab:adapt}
    \begin{tabular}{c|c|c}
    \hline
    Method & sim novel & real novel\\
    \hline
    ClassIndis & 56.8 & 37.3\\
    EncoderIndis & 72.0 & 60.3\\
    NoMetric & 68.1 & 53.6\\
    
    One-Grasp (\cite{yang2021attribute}) & 83.7 & 76.6\\
    AttrID-One-Grasp & 79.7 & 73.0\\
    Adversarial+One-Grasp (Ours) & \textbf{86.0} & \textbf{81.7}\\
    \hline
    \end{tabular}
\end{table}
\begin{figure}[!t]
  \centering
  \includegraphics[width=0.485\textwidth]{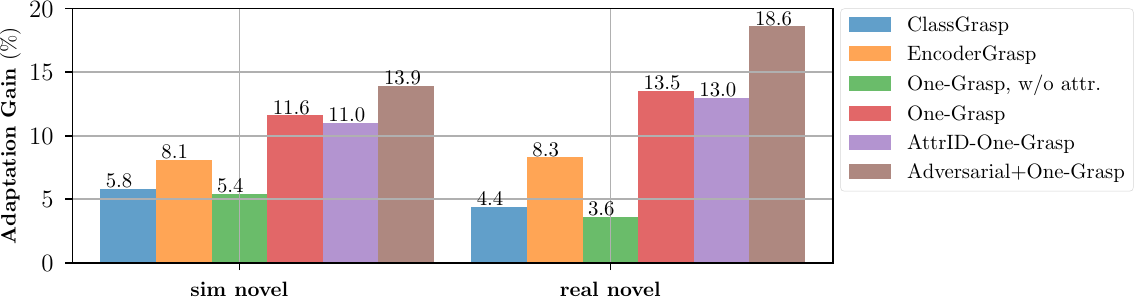}
  \caption{\textbf{The adaptation gains} of instance grasping made by different adaptation methods. The plot shows the effectiveness of our \textit{One-Grasp} and \textit{Adversarial+One-Grasp} adaptation approach, which achieve significant adaptation gains.}
  \label{fig:adapt_gain}
\end{figure}

We also update \textit{Indiscriminative} grasping in \textit{ClassIndis} and \textit{EncoderIndis}. We keep the experimental setup the same with Sec. \ref{subsec:exp_grasp} and evaluate the instance grasping performance of the adapted models. We report the adapted instance grasping success rate in Table \ref{tab:adapt} and the adaptation gains in Fig. \ref{fig:adapt_gain}. The attributes classifier in \textit{ClassIndis} suffers from insufficient adaptation data, limiting its target recognition and adaptation performance. While \textit{EncoderIndis} minimizes embedding distances in its latent space and shows better performance, it is still worse than \textit{One-Grasp}. By fine-tuning over the structured metric space, \textit{One-Grasp} updates the end-to-end model and improves target recognition and grasping jointly. At the cost of minimal data collection, \textit{One-Grasp} achieves an 83.7\% grasping success rate on the simulated novel objects and an 76.6\% grasping success rate on the real objects, which shows the significant adaptation gains. On the contrary, the unstructured latent space in \textit{NoMetric} limits its adaptation, demonstrating the importance of attributes learning for grasping affordances learning. The difference in adaptation performance between \textit{NoMetric} and \textit{One-Grasp} demonstrates the significance of the regulated feature space in adaptation. Furthermore, the substantial adaptation gain observed in \textit{AttrID} validates the applicability of our adaptation method with sparse feature inputs.

\begin{figure}[!t]
    \centering
    \begin{subfigure}{0.48\textwidth}
        {\includegraphics[width=0.24\textwidth]{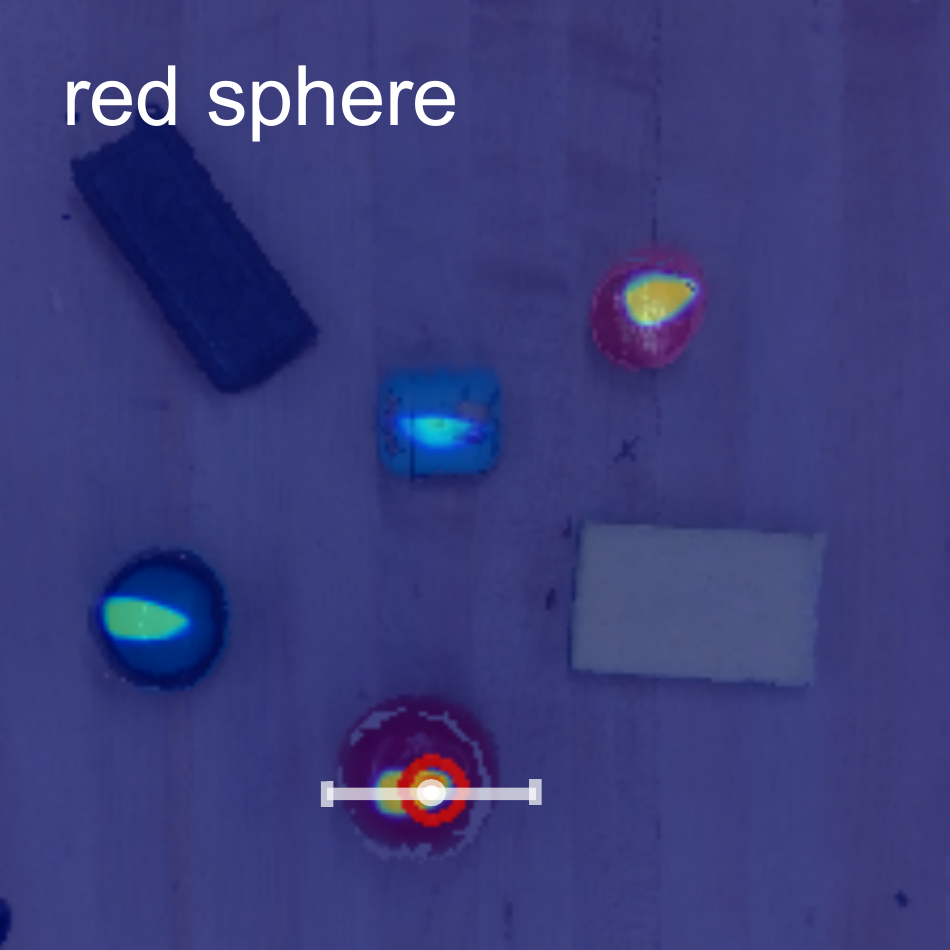}}\hfill
        {\includegraphics[width=0.24\textwidth]{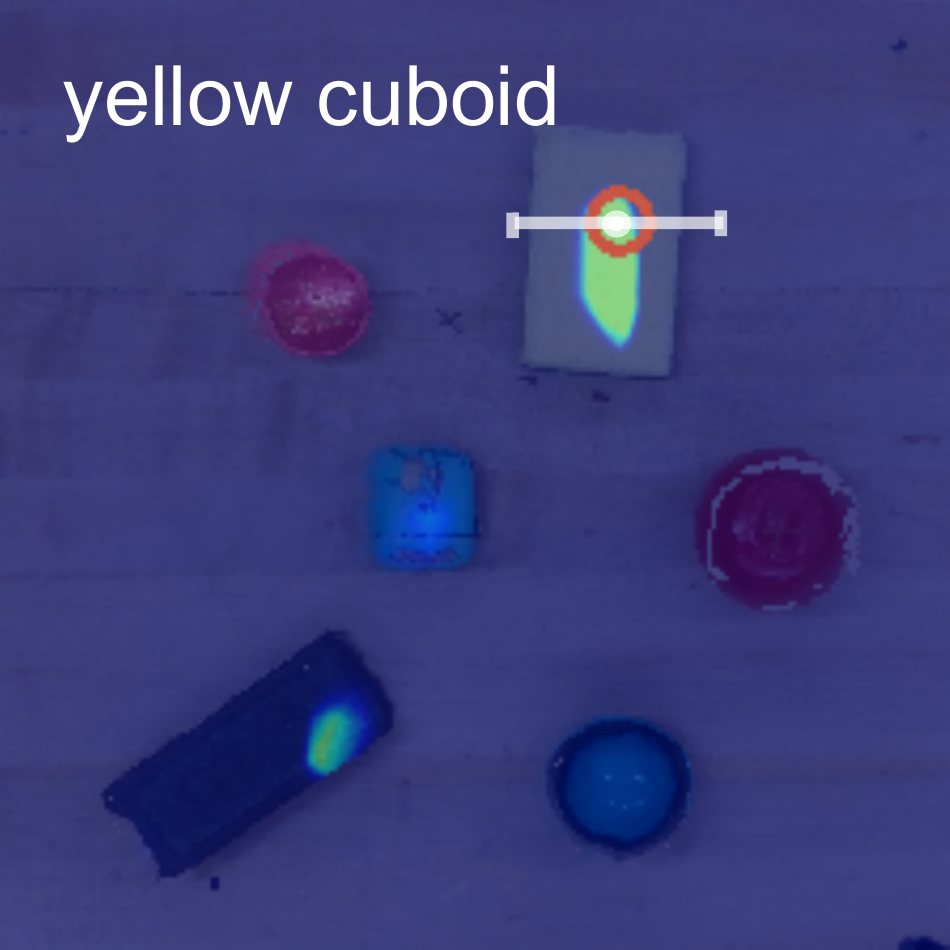}}\hfill
        {\includegraphics[width=0.24\textwidth]{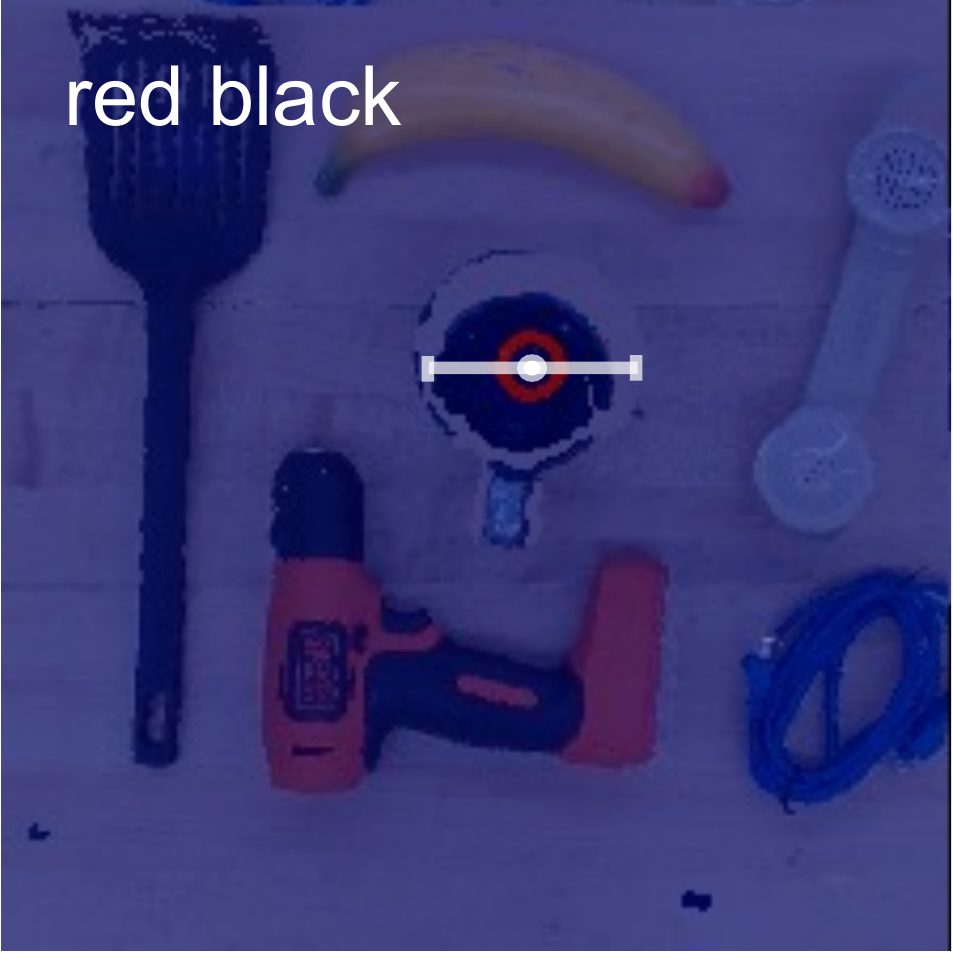}}\hfill
        {\includegraphics[width=0.24\textwidth]{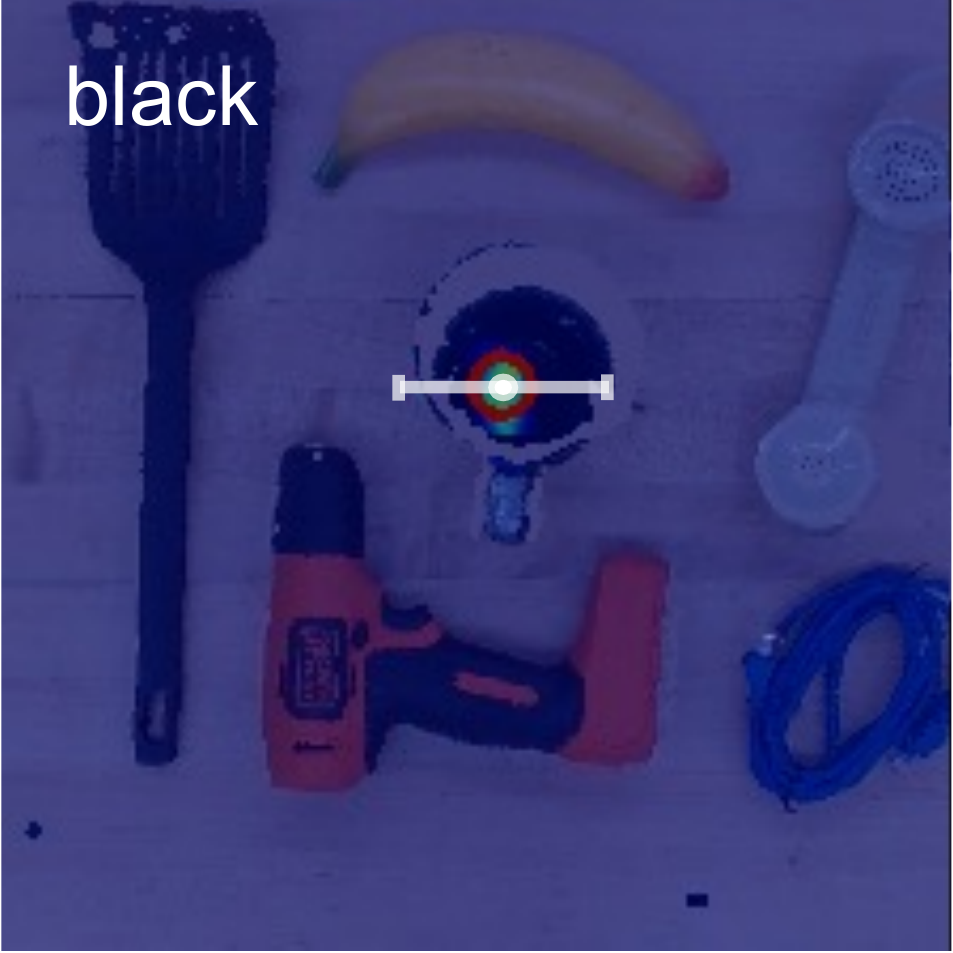}}%
        \vspace{-3pt}
        \caption{\textit{Generic} model before adaptation}
        \label{fig:generic_no_adv}
        \vspace{5pt}
    \end{subfigure}
    \begin{subfigure}{0.48\textwidth}
        {\includegraphics[width=0.24\textwidth]{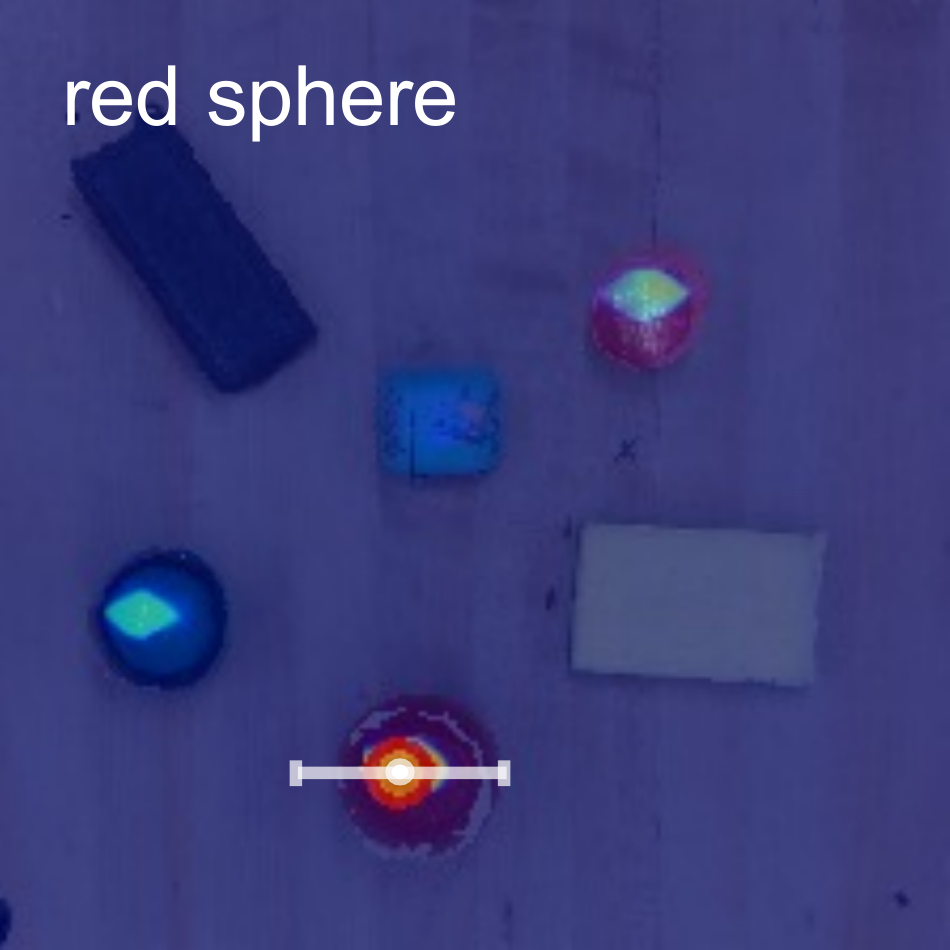}}\hfill
        {\includegraphics[width=0.24\textwidth]{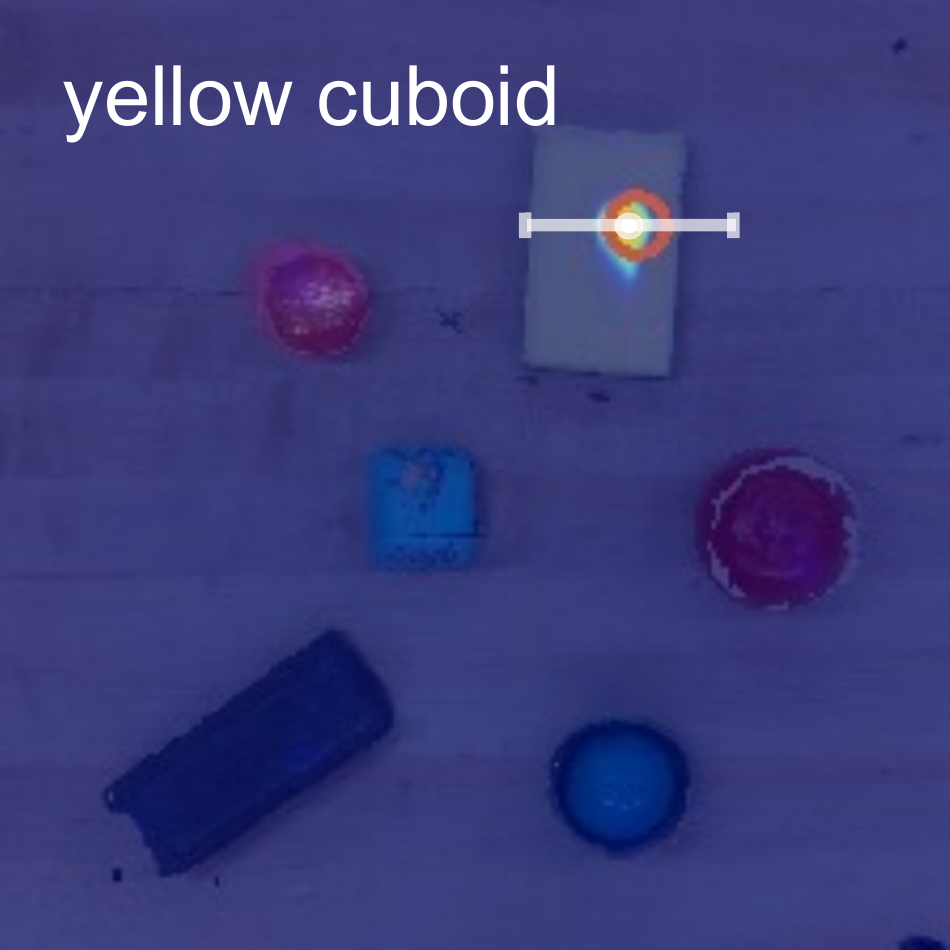}}\hfill
        {\includegraphics[width=0.24\textwidth]{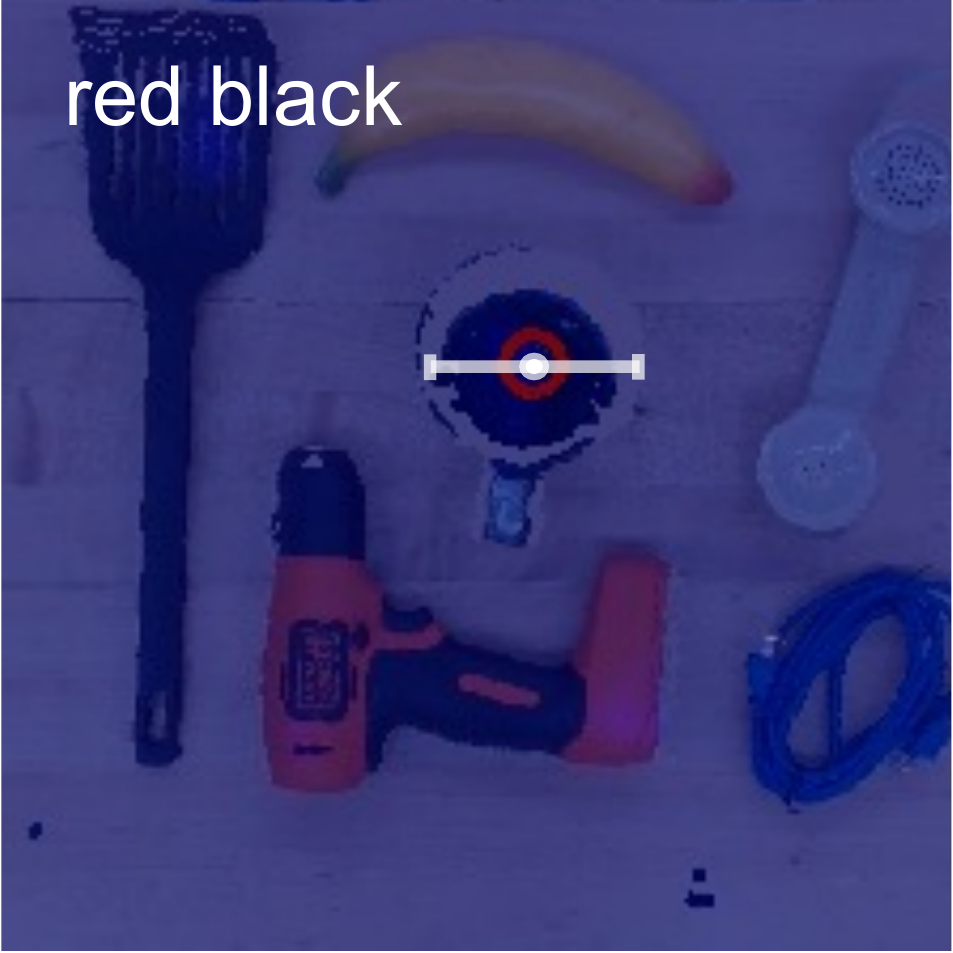}}\hfill
        {\includegraphics[width=0.24\textwidth]{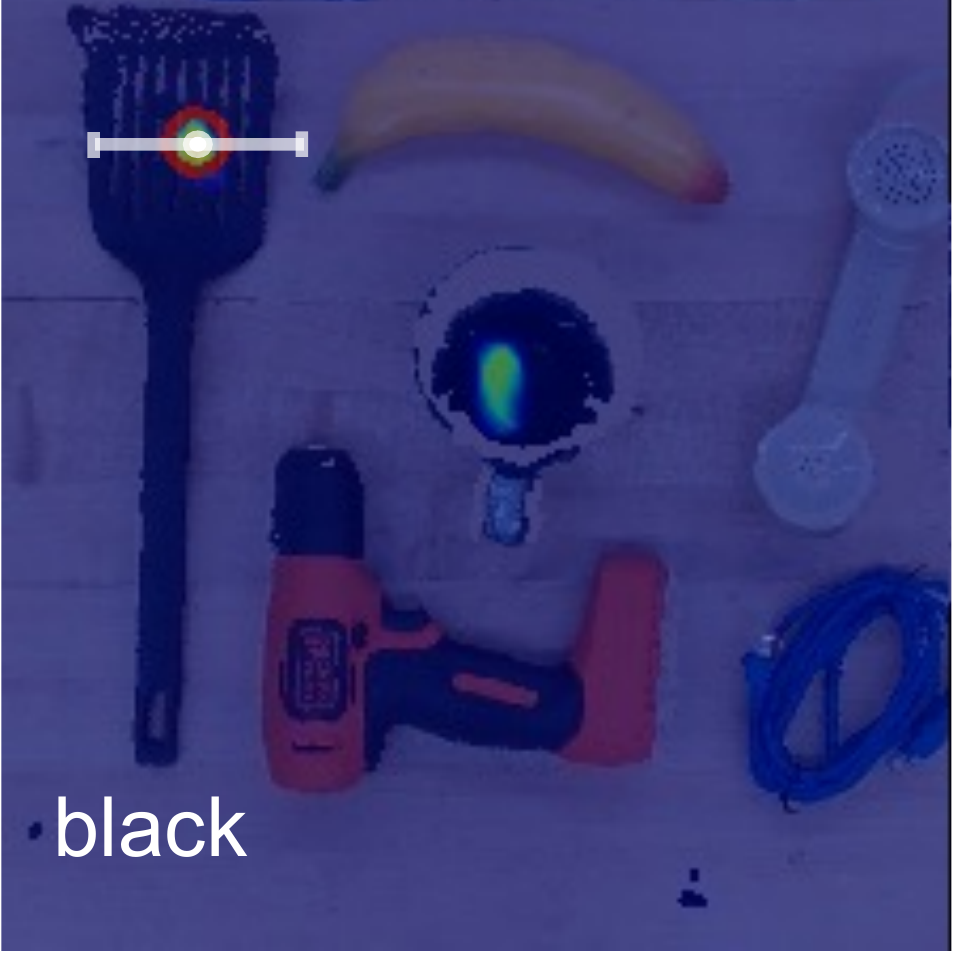}}%
        \vspace{-3pt}
        \caption{\textit{Adversarial} adaptation}
        \label{fig:real_adv}
        \vspace{5pt}
    \end{subfigure}
    \begin{subfigure}{0.48\textwidth}
        {\includegraphics[width=0.24\textwidth]{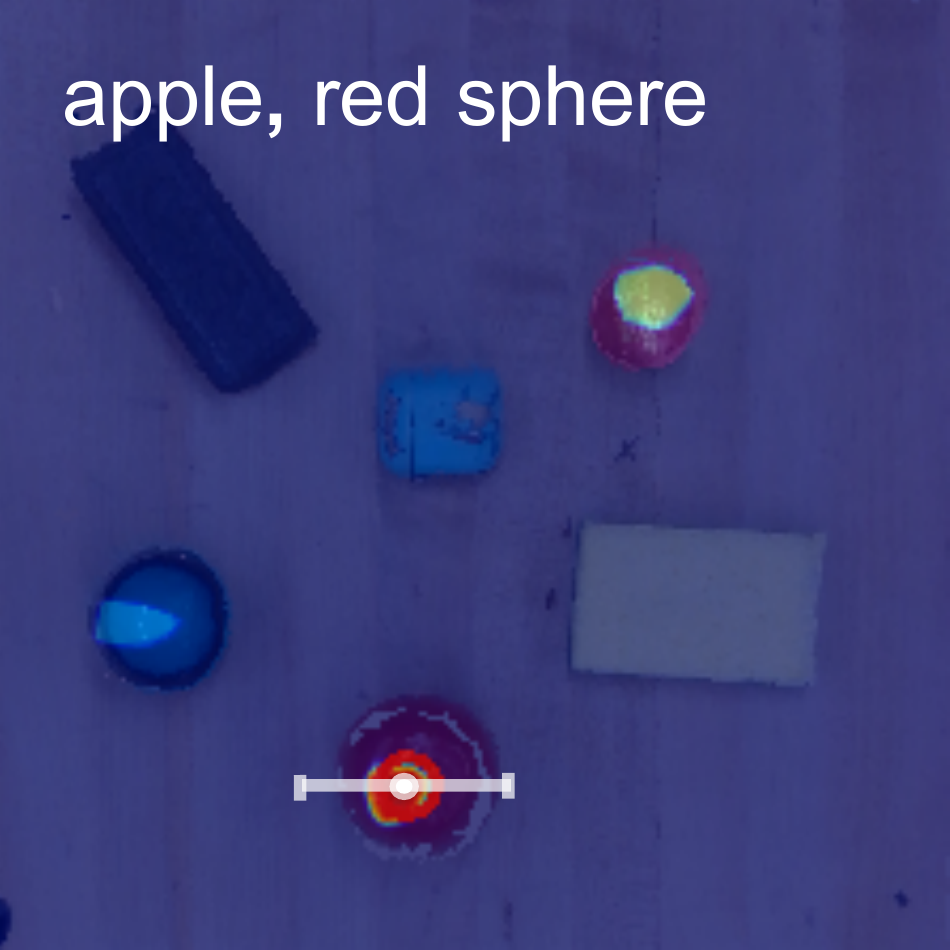}}\hfill
        {\includegraphics[width=0.24\textwidth]{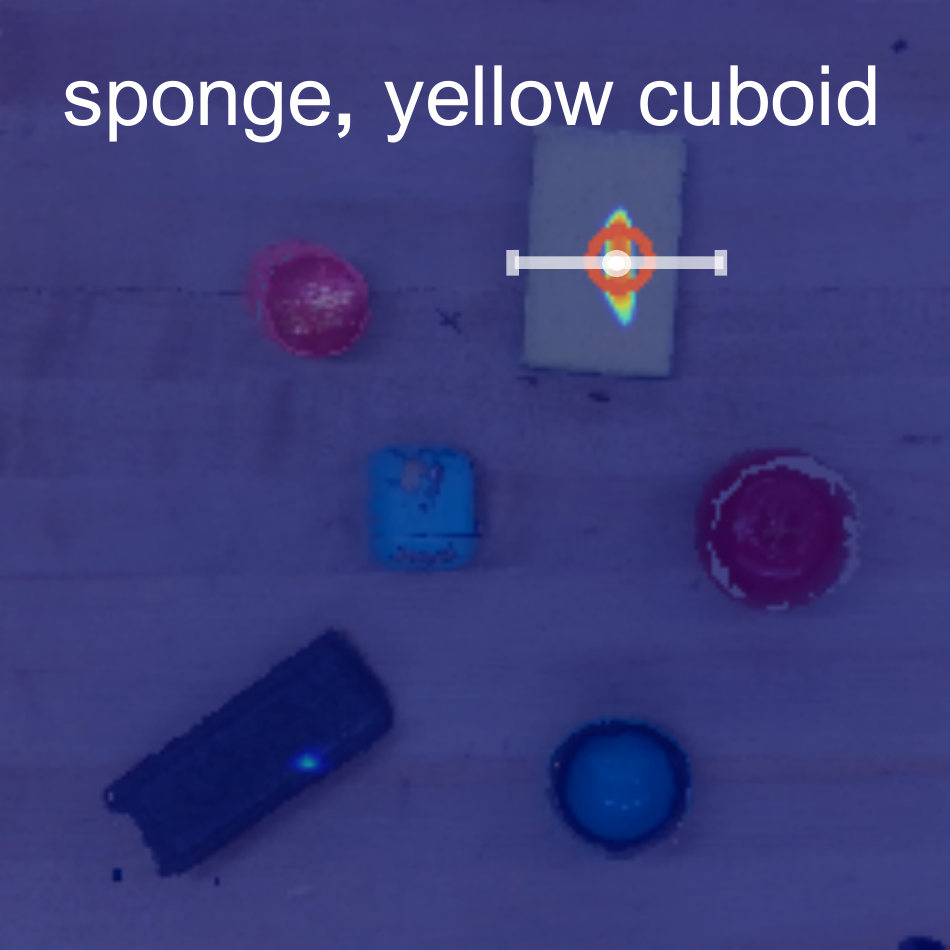}}\hfill
        {\includegraphics[width=0.24\textwidth]{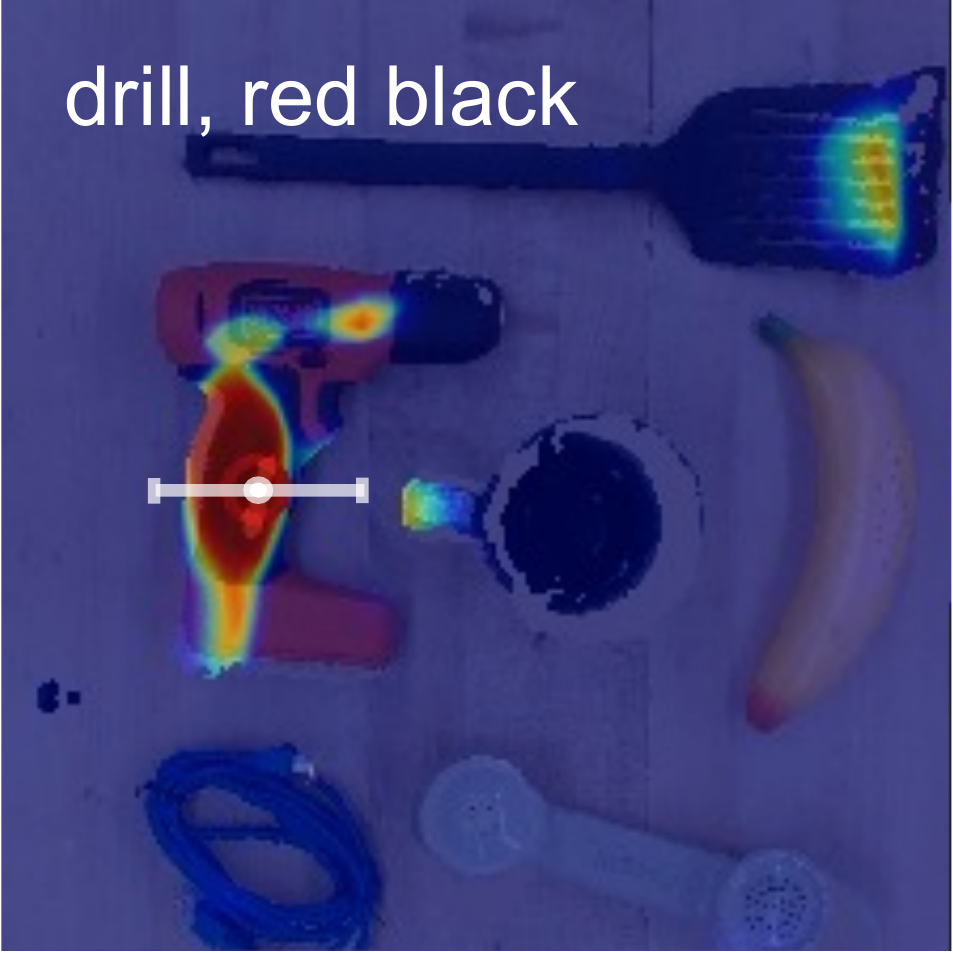}}\hfill
        {\includegraphics[width=0.24\textwidth]{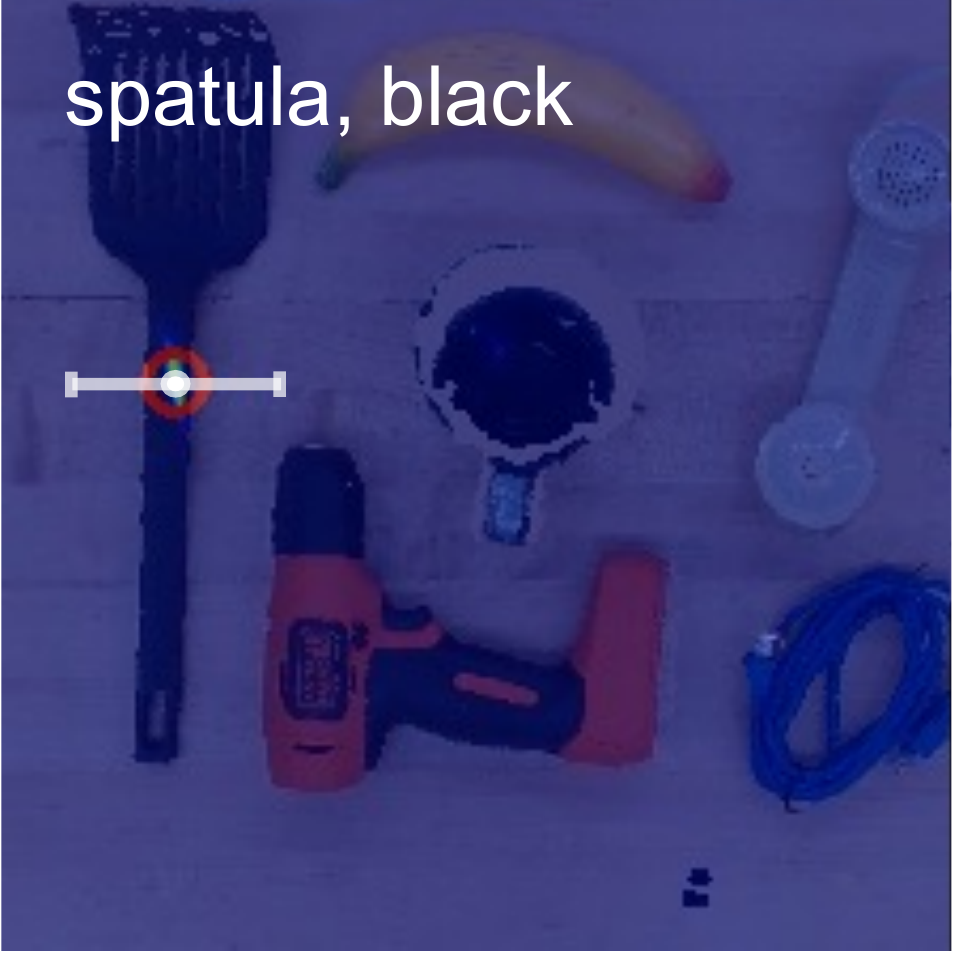}}%
        \vspace{-3pt}
        \caption{\textit{One-Grasp} adaptation}
        \label{fig:real_1grasp}
        \vspace{5pt}
    \end{subfigure}
    \begin{subfigure}{0.48\textwidth}
        {\includegraphics[width=0.24\textwidth]{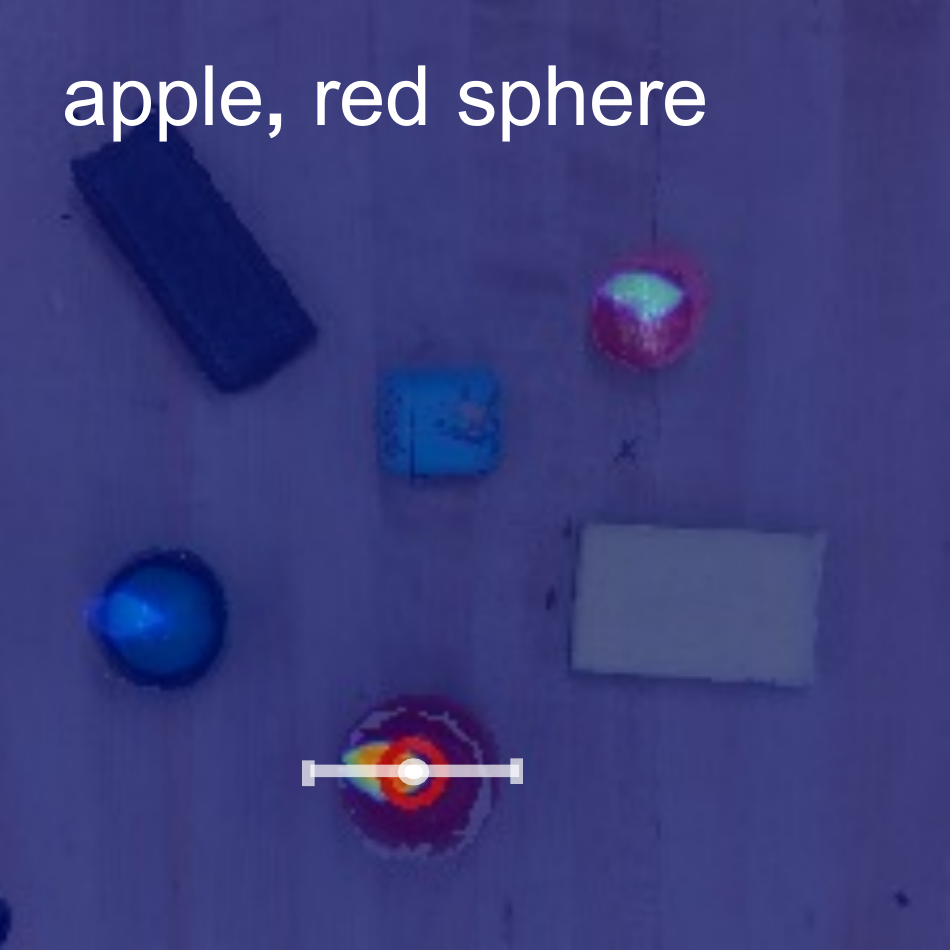}}\hfill
        {\includegraphics[width=0.24\textwidth]{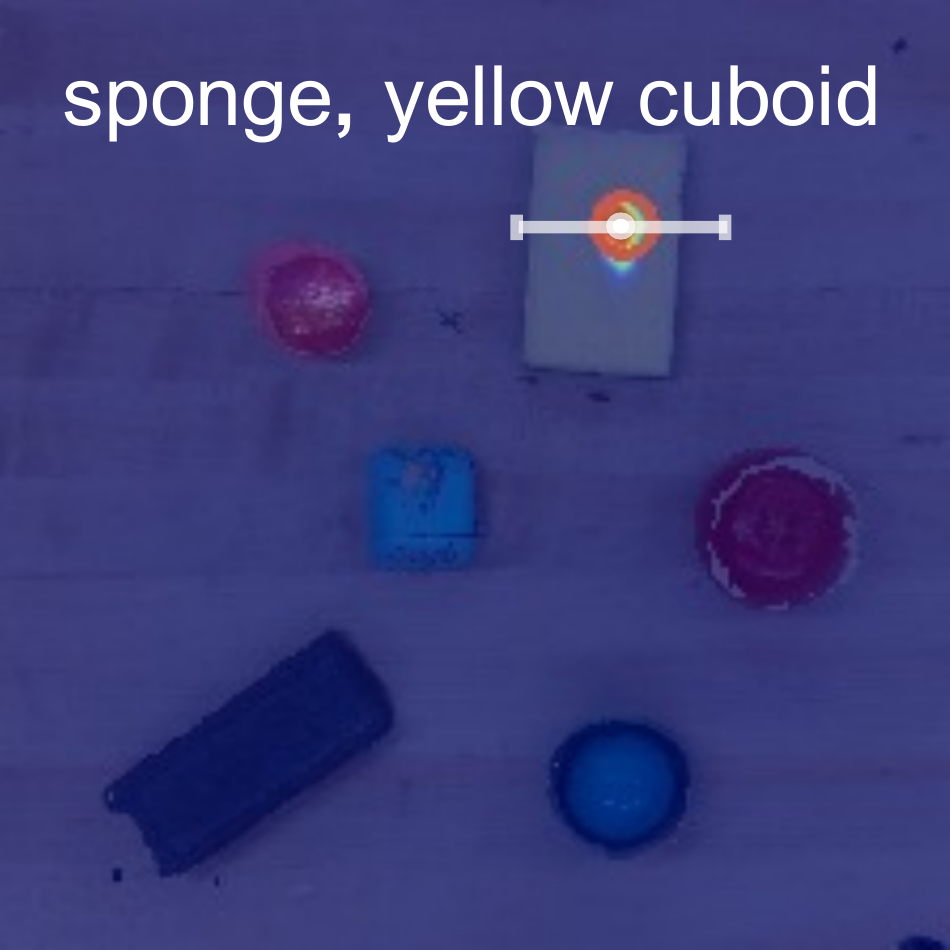}}\hfill
        {\includegraphics[width=0.24\textwidth]{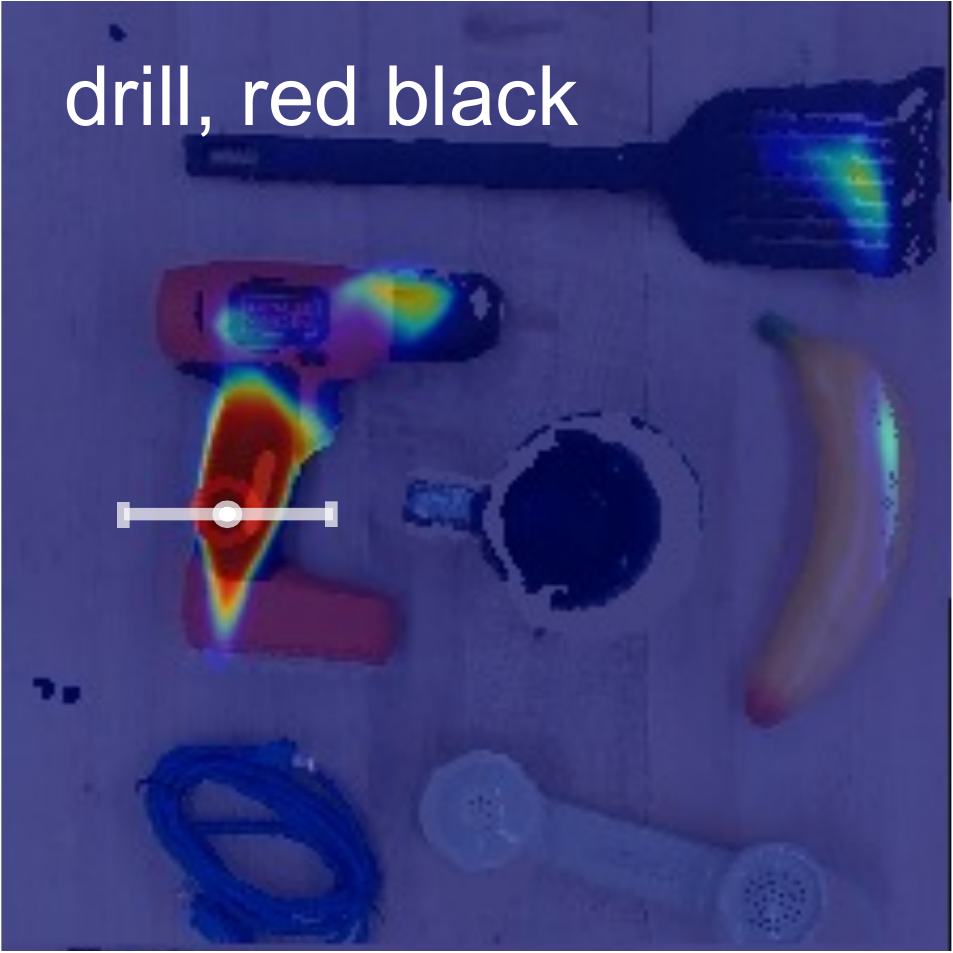}}\hfill
        {\includegraphics[width=0.24\textwidth]{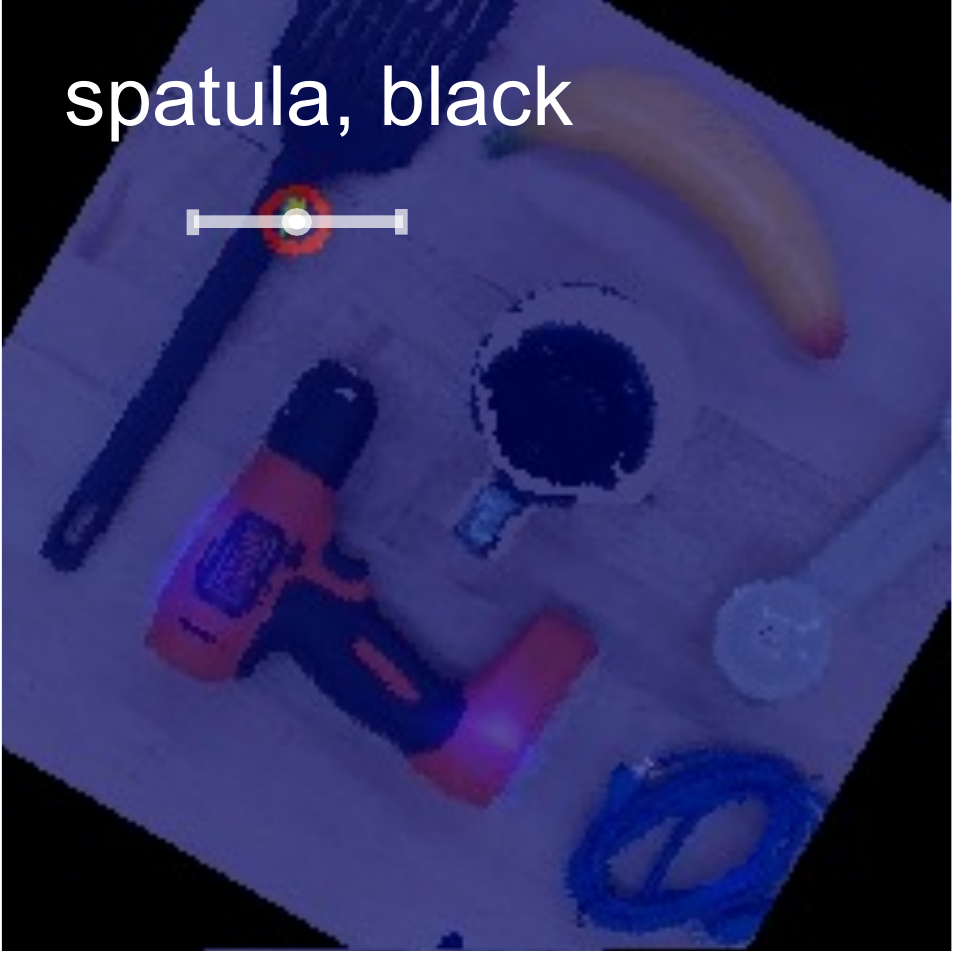}}%
        \caption{\textit{Adversarial+One-Grasp} adaptation}
        \label{fig:adv_and_1grasp}
    \end{subfigure}
    \caption{\textbf{Visualization of grasping maps before and after adaptation.} (\subref{fig:generic_no_adv}) shows the grasping affordances from the generic model trained only with simulated basic objects, and (\subref{fig:real_adv}) to (\subref{fig:adv_and_1grasp}) show the affordances from the adapted models after \textit{Adversarial}, \textit{One-Grasp}, and \textit{Adversarial+One-Grasp} adaptation, respectively. 
    Our adaptation methods, which require only a limited amount of adaptation data, effectively enhance model performance.
    }
    \label{fig:real}
\end{figure}

As an improvement, we propose applying \textit{Adversarial} adaptation on the image encoder to learn domain-invariant features before \textit{One-Grasp} adaptation. The image encoder's domain invariancy results in superior transferring performance for \textit{Adversarial+}\textit{One-Grasp}: an instance grasping success rate of 86.0\% in the domain of sim novel and 81.7\% in the domain of real-world novel (real novel). The qualitative results in Fig. \ref{fig:real} suggest the efficacy of the two adaptation methods: 1) both \textit{Adversarial} adaptation and \textit{One-Grasp} adaptation increase the recognition and grasping performance of the models, and 2) \textit{Adversarial} adaptation minimizes grasping noises around non-target objects (by reducing domain feature changes), while \textit{One-Grasp} adaptation can rectify recognition and grasping errors through end-to-end updates. The more compact and centered contour in Fig. \ref{fig:real_adv} could be explained by the hypothesis that domain-invariant features improve the output consistency of the encoder across domains. The corrected target recognition in Fig. \ref{fig:real_1grasp}, on the other hand, is attributed to the \textit{One-Grasp} adaptation which effectively shifts the affordances from irrelevant objects to the target object through the end-to-end model updates. Another noteworthy finding is that the combined adaptation \textit{Adversarial+One-Grasp} appears to benefit from both \textit{Adversarial} adaptation and \textit{One-Grasp} adaptation, as their focuses are complementary.

\begin{figure}[!t]
  \centering
  \begin{subfigure}{0.48\textwidth}
    \includegraphics[width=\textwidth]{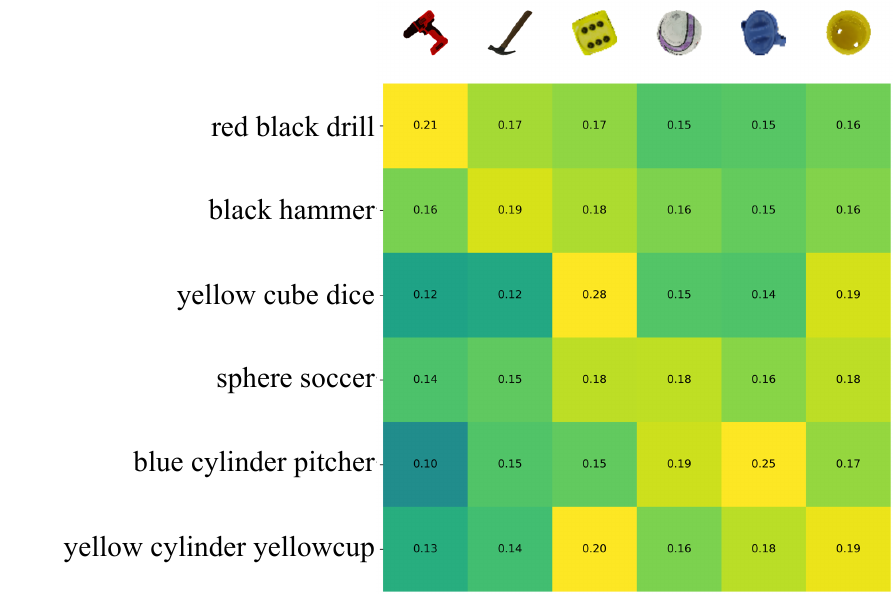}
    \caption{CLIP matching scores}
    \label{fig:confusion0}
  \end{subfigure}
  \vspace{3mm}
  \begin{subfigure}{0.48\textwidth}
    \includegraphics[width=\textwidth]{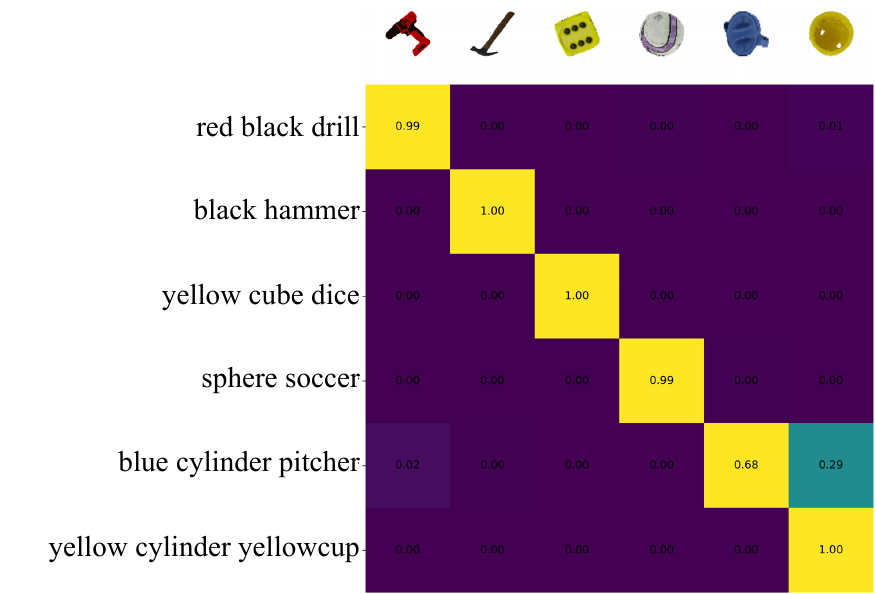}
    \caption{\textit{Adversarial+One-Grasp} scores}
    \label{fig:confusion1}
  \end{subfigure}
  \caption{\textbf{The visualization of confusion matrices.} The comparison between the two confusion matrices shows the effectiveness of our adaptation method trained with a minimum amount of data.}
  \label{fig:confusion}
\end{figure}

\subsection{Comparison with a Foundation Model}

In addition to the instance grasping experiments, we perform another quantitative evaluation of the adapted model on sim novel objects. For comparison, we utilize the CLIP model \cite{radford2021learning}, a recently prevailing multimodal (text and image) foundation model. CLIP aligns language and image features through training with millions of text-image pairs and is widely acclaimed for its robust generalization capability across various objects. To evaluate the performance of CLIP, we segment and crop randomly placed objects in the workspace. For each object crop and its attribute text description, the CLIP model calculates a matching score, which is then used to construct the confusion matrix shown in Fig. \ref{fig:confusion0}. The pairwise matching scores reflect the similarity between the object crops and the text inputs. As for our \textit{Adversarial+One-Grasp} model, we compute another confusion matrix shown in Fig. \ref{fig:confusion1} by collecting the highest affordance value, under each target attribute description, from affordance maps $Q_g$ within each object's segmentation mask. While CLIP matching uses cropped images of single objects, \textit{Adversarial+One-Grasp} is tested using workspace images of multiple objects, which is a more natural but also more challenging setting. Fig. \ref{fig:confusion} shows that \textit{Adversarial+One-Grasp} achieves more accurate grounding of the correct target objects. In contrast, the matching scores produced by the CLIP model often lack discrimination, leading to some misclassifications (e.g., ``sphere soccer'' and ``yellow cylinder yellowcup''). This comparison highlights the limitations of zero-shot generalization in a foundation model and showcases the effectiveness of our adaptation method.

\subsection{Ablative Analysis of Adaptation} 

\begin{figure*}[!t]
  \centering
  \includegraphics[width=\textwidth]{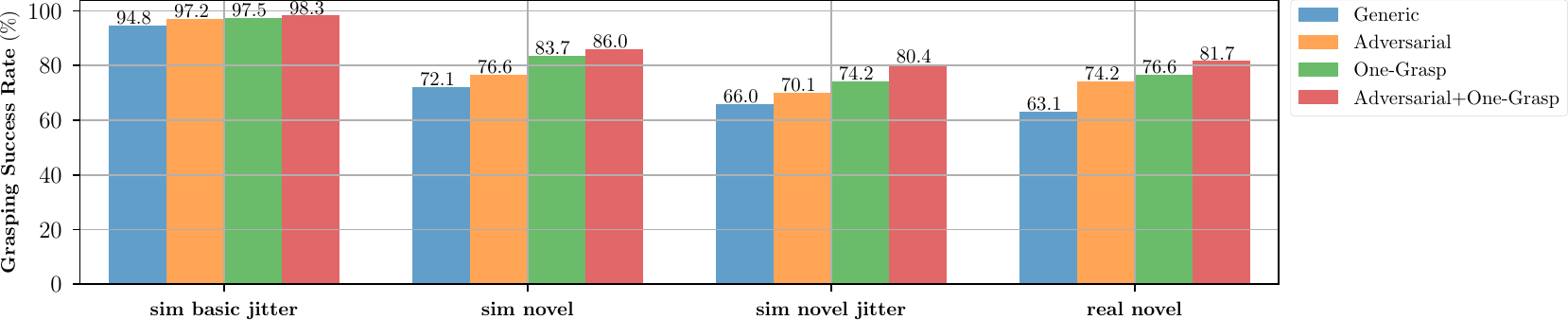}
  \caption{\textbf{Adaptation performance in the designed testing environments.} The instance grasping success rate of four approaches in the four testing environments. The plot shows the effectiveness of the proposed two data-efficient adaptation methods, \textit{Adversarial} and \textit{One-Grasp} adaption, which consistently improve the grasping performance. It is worth noting the additive performance of the two methods, and the combined \textit{Adversarial}+\textit{One-Grasp} adaption performs the best in all the testing cases.}
  \label{fig:adapt_comp}
\end{figure*}
\begin{figure}[!t]
    \centering
    \begin{subfigure}[t]{0.1152\textwidth}
        \includegraphics[width=\textwidth]{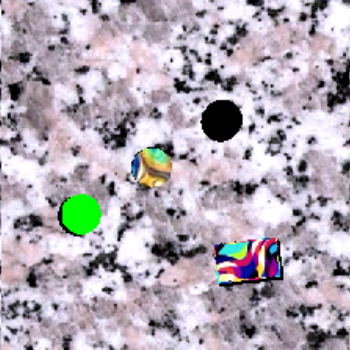}
        \caption*{sim basic jitter}
    \end{subfigure}
    \begin{subfigure}[t]{0.1152\textwidth}
        \includegraphics[width=\textwidth]{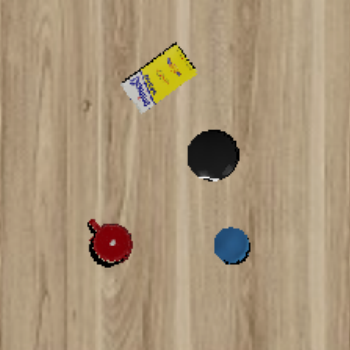}
        \caption*{sim novel}
    \end{subfigure}
    \begin{subfigure}[t]{0.1152\textwidth}
        \includegraphics[width=\textwidth]{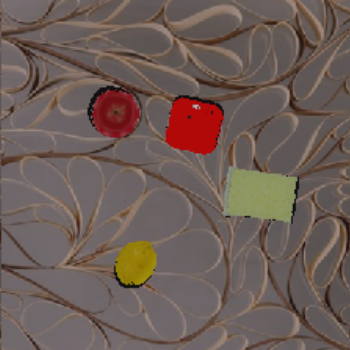}
        \caption*{sim novel jitter}
    \end{subfigure}
    \begin{subfigure}[t]{0.1152\textwidth}
        \includegraphics[width=\textwidth]{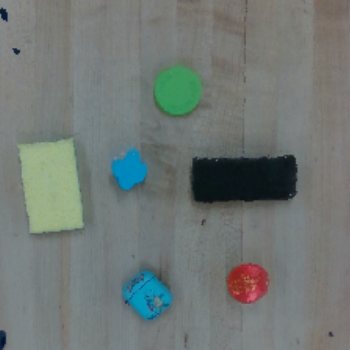}
        \caption*{real novel}
    \end{subfigure}
    \caption{\textbf{Example images} of the designed testing environments for ablative analysis of adaptation. The environments are created to test the adaption performance of the grasping models as the degree of domain shift increases.}
    \label{fig:adapt_env}
\end{figure}

\label{subsec:abla}
The proposed adaptation approaches efficiently improve the instance grasping performance of our model. As discussed in Sec. \ref{subsec:exp_adapt}, the finding that the approaches have complementary adaptation focuses leads to one of the major features: the two adaptation methods can be employed individually or in combination, depending on the availability of adaptation data. To investigate their independent and combinative performance, we conduct an ablation study to compare the grasping models as follows:
\begin{itemize}[noitemsep, topsep=0pt]
    \item[1)] \textit{Generic} is the baseline generic model obtained in Sec. \ref{subsec:exp_grasp} before any adaptation.
    \item[2)] \textit{Adversarial} adapts the generic model to learn domain-invariant features using a large augmented data, as discussed in Sec. \ref{subsec:method_adv}.
    \item[3)] \textit{One-Grasp} adapts the generic model using one grasping trial of the target object, as discussed in Sec. \ref{subsec:method_1grasp}.
    \item[4)] \textit{Adversarial+One-Grasp} applies the two adaptation approaches on the generic model sequentially to improve the recognition and grasping.
\end{itemize}
As shown in Fig. \ref{fig:adapt_env}, the adaptation methods are compared in four testing environments with an increasing extent of domain shifts: 1) sim basic jitter---simulated basic objects with visual jitter (see Sec. \ref{subsec:method_syssim}) applied on color and depth channels as well as background, 2) sim novel---simulated novel objects, 3) sim novel jitter---simulated novel objects with visual jitter, and 4) real novel---real novel objects. As the training environment uses simulated basic objects, the testing environments include the domain shifts caused by novel objects, visual jitter, and real-world noises.

We report the experimental results of the instance grasping success rate in Fig. \ref{fig:adapt_comp}. Overall, all adaptation methods improve the grasping performance across the testing environments. It is not surprising that adaptation is likely to be more effective (i.e., leading to more performance increments) if the domain shifts are severer. For example, the adaptation gain in the environment of sim basic jitter is less than 4\%, while the real environment witnesses an adaptation gain of over 18\%. Moreover, when encountering complex novel objects that are more challenging (e.g., drill) than the basic training objects (e.g., red cuboid), the \textit{One-Grasp} method provides more adaptation power than the \textit{Adversarial} method. In \textit{Adversarial} adaptation, we use unlabeled data from the target domain to update the image encoder, while the text encoder and grasping decoder remain unadapted. As a result, the \textit{Adversarial} adapted model is more prone to encountering difficulties with challenging novel objects (e.g., drill and spatula). On the other hand, \textit{One-Grasp} adaptation adapts the entire model and demonstrates better performance on these challenging objects, but it requires additional labeled data (at least one grasp) of the objects. Another observation is that we can combine the two adaptation methods to achieve an even higher adaptation performance. Through various tests, the combined method \textit{Adversarial+One-Grasp} adaptation consistently shows the best performance in all the testing environments. This suggests that the two adaptation methods adapt our model complementarily for accumulative adaptation gains. In practice, we can choose a configuration of adaptation methods based on the availability of the adaptation data.

\subsection{Data Augmentation for Adaptation}
%
%

\begin{table}[!t]
    \centering
    \caption{Ablations for Object-Level Augmentation (\%)}
    \label{tab:abl_adv}
    \begin{tabular}{c|c|c}
    \hline
    Method & sim novel & real novel\\
    \hline
    Objects (w/o overlay and single-object aug.) & 73.8 & 70.0\\
    ObjectOverlay (w/o single-object aug.) & 76.1 & 72.6\\
    ObjectAug (Ours) & \textbf{76.6} & \textbf{74.2}\\
    \hline
    \end{tabular}
    \vspace{13pt}
    \centering
    \caption{Ablations for One-Grasp Augmentation (\%)}
    \label{tab:abl_1grasp}
    \begin{tabular}{c|c|c}
    \hline
    Method & sim novel & real novel\\
    \hline
    OneGrasp (w/o repetition and rotation) & 81.5 & 72.6\\
    OneGraspRpt (w/o rotation) & 81.7 & 73.8\\
    OneGraspAug (Ours) & \textbf{83.7} & \textbf{76.6}\\
    \hline
    \end{tabular}
\end{table}

The quality of adaptation data is critical for grasping adaptation. The two data augmentation methods, \textit{ObjectAug} and \textit{OneGraspAug}, are proposed for the two adaptation methods respectively. We execute the ablation studies in Table \ref{tab:abl_adv} and Table \ref{tab:abl_1grasp} to examine the augmentation methods. The results of adapted instance grasping are presented in the tables. In the ablations for object-level augmentation, the compared approaches are
\begin{itemize}[noitemsep, topsep=0pt]
    \item[1)] \textit{Objects} uses the raw data (images of single objects) as the adaptation data.
    \item[2)] \textit{ObjectOverlay} overlays the randomly sampled objects on the background image to synthesize a large dataset covering possible object combinations and locations.
    \item[3)] \textit{ObjectAug} is our object-level data augmetation method discussed in Sec. \ref{subsec:method_adv}, where much richer object configurations (i.e., orientations and scales) are covered in the synthesized dataset.
\end{itemize}
For the above augmentation methods, we keep the dataset size constant and use each augmentation data in (\ref{eq:adva}), accordingly. Even though the data is unlabeled, the adaptation data quality has a direct impact on grasping performance, as seen in Table \ref{tab:abl_adv}. The performance difference between \textit{Objects} and \textit{ObjectAug}, for example, is up to 4\% on real novel objects, despite the fact that they nominally contain the identical objects. This finding demonstrates that the suggested object-level data augmentation successfully reduces domain shift by supplying rich unlabeled data.

In the ablations for one-grasp augmentation, the comparable approaches include
\begin{itemize}[noitemsep, topsep=0pt]
    \item[1)] \textit{OneGrasp} uses the raw data of one successful grasp trial, including an RGB-D image and the corresponding grasping action.
    \item[2)] \textit{OneGraspRpt} simply repeats the one-grasp data $N$ times without rotating the data, where $N=6$ is the angle discretion parameter.
    \item[3)] \textit{OneGraspAug} is our one-grasp data augmetation method discussed in Sec. \ref{subsec:method_1grasp}, where we augment the one-grasp data by rotating for $N$ orientations to enrich possible orientations of objects and robot grasping.
\end{itemize}
As shown in Table \ref{tab:abl_1grasp}, \textit{OneGraspAug} outperforms the compared methods by over 4\% grasping success rate on real robot experiments, which demonstrates how angle-augmented data can be used to make the grasping model rotation-invariant for object recognition and grasping.

\section{Conclusion}
In this work, we presented a novel attribute-based robotic grasping system. An end-to-end architecture was proposed to learn object attributes and manipulation jointly. Workspace images and query text were encoded into a joint metric space, which was further supervised by object persistence before and after grasping. Our model was self-supervised in a simulation only using basic objects but showed good generalization. To further adapt to novel objects and real-world scenes, we proposed two data-efficient adaptation methods, adversarial adaptation and one-grasp adaptation, which only require unlabeled object images or one grasp trial. Our grasping system achieved an $86.0\%$ instance grasping success rate in simulation and an $81.7\%$ instance grasping success rate in the real world, both on unknown objects. Our approach outperformed the other compared methods by large margins. 

We showed that incorporating object attributes in robotic grasping improves the performance of the deep learning grasping model. To the best of our knowledge, this is the first work to explore object attributes for the generalization and adaptation of deep learning robotic grasping models. Our long-term goal is to further improve the effectiveness and robustness of our model by pre-training it with objects of richer attributes. Another possible avenue for future work is to study object attributes under partial observation, such as the shape of an object in dense clutter. It would be of interest to explore how to achieve more robust attribute perception, potentially using shape completion algorithms. 

\bibliographystyle{IEEEtran}
\bibliography{references}

\newpage
\begin{IEEEbiography}[{\includegraphics[width=1in,height=1.25in,clip,keepaspectratio]{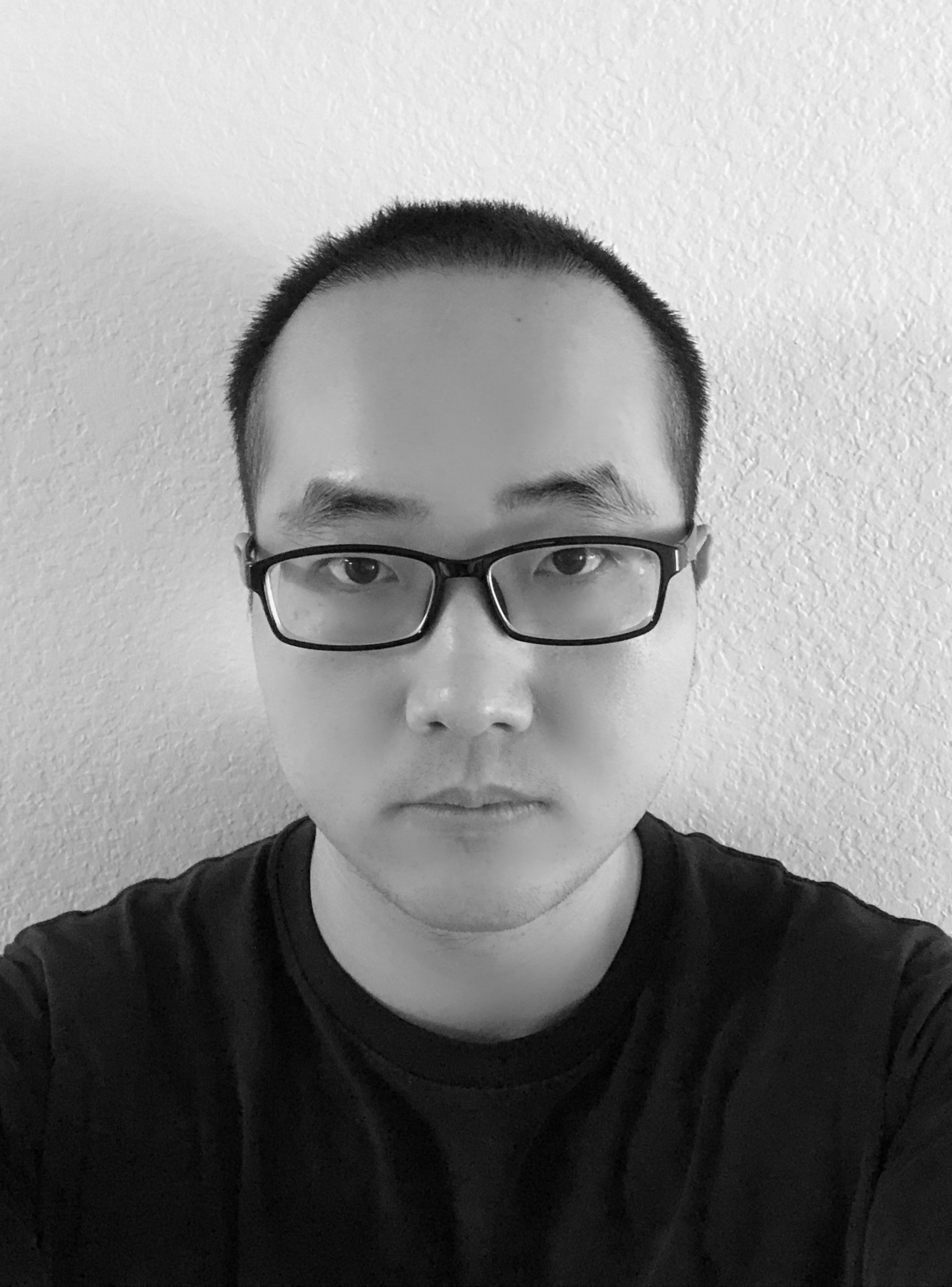}}]{Yang Yang}
received the B.Eng degree in Energy and Power Engineering from Huazhong University of Science and Technology, Wuhan, China, in 2015, and the Ph.D. degree in Computer Science from University of Minnesota, Minneapolis, USA, in 2022. He joined Meta shortly before graduation, and he is currently a Senior Research Scientist with Meta AI, working on ultra-scale ranking and foundation models.

Dr. Yang is the recipient of the ICRA 2022 Outstanding Student Paper Award and the UMN 2021 UMII-MnDRIVE Graduate Assistantship. His research interests include recommendation and monetization AI, embodied AI, and AGI.
\end{IEEEbiography}
\vspace*{-5\baselineskip}

\begin{IEEEbiography}[{\includegraphics[width=1in,height=1.25in,clip,keepaspectratio]{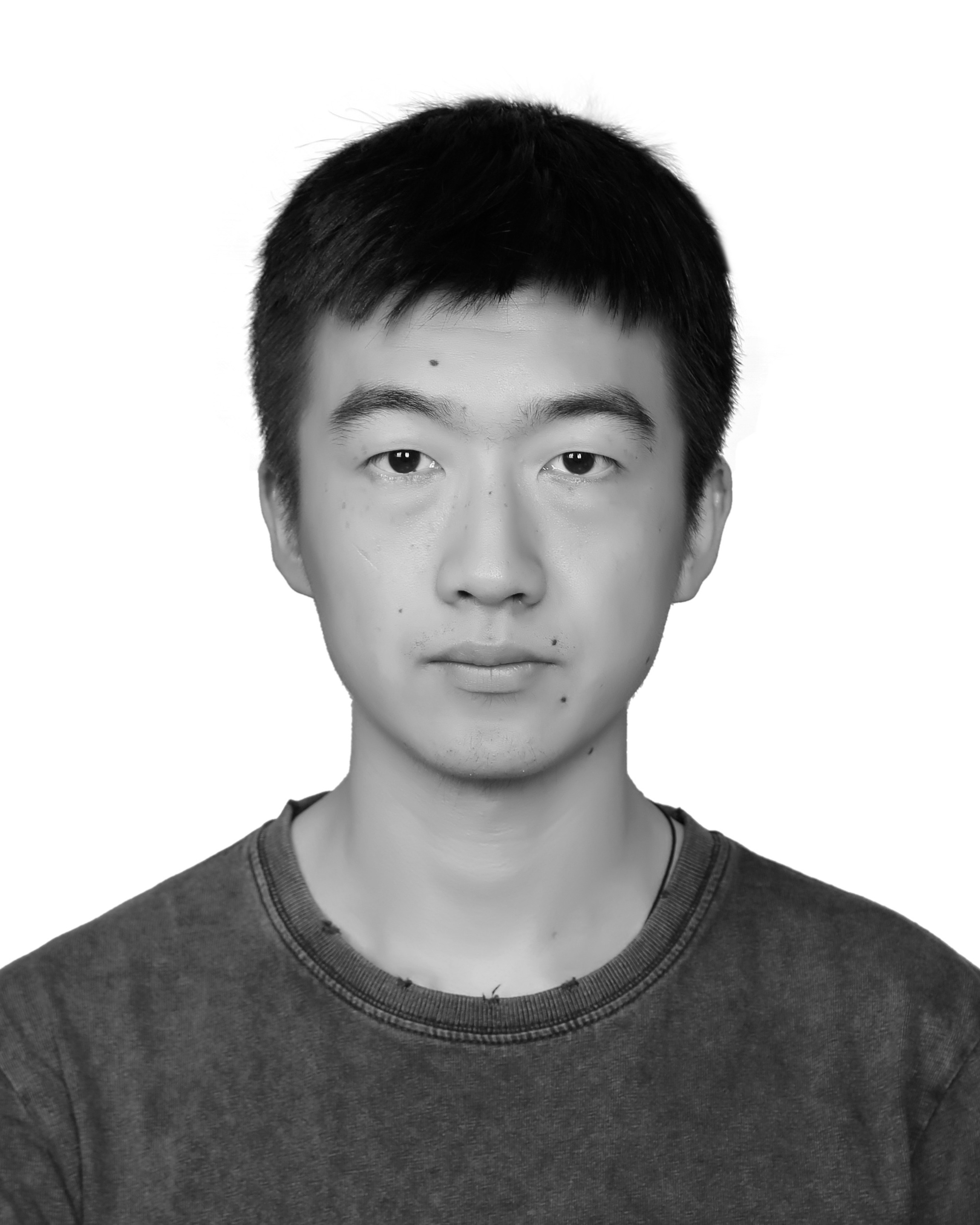}}]{Houjian Yu} received the B.Eng degree in Electrical Engineering from North China Electric Power University, Beijing, China, in 2018, and the M.S. degree in Electrical and Computer Engineering from University of California San Diego, USA, in 2020. He is now a Computer Engineering Ph.D. candidate at the University of Minnesota, Minneapolis, USA. His research interests are in robotics and machine learning.
\end{IEEEbiography}
\vspace*{-5\baselineskip}

\begin{IEEEbiography}[{\includegraphics[width=1in,height=1.25in,clip,keepaspectratio]{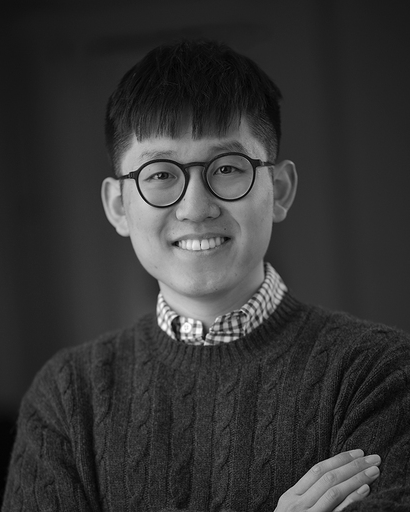}}]{Xibai Lou} received the B.Eng and M.S. degrees in Electrical Engineering from Georgia Institute of Technology, Atlanta, USA, in 2014 and 2015, respectively, and the Ph.D. degree in Electrical Engineering from University of Minnesota, Minneapolis, USA, in 2023. He is currently an Applied Scientist with Amazon Robotics, Seattle, USA, focusing on deep learning in 3D robot vision for warehouse automation.

Dr. Lou is the recipient of the ICRA 2022 Outstanding Student Paper Award and the UMN 2022 UMII-MnDRIVE Graduate Assistantship. His research interests include robot manipulation, perception, and deep learning-based algorithms.
\end{IEEEbiography}
\vspace*{-5\baselineskip}

\begin{IEEEbiography}[{\includegraphics[width=1in,height=1.25in,clip,keepaspectratio]{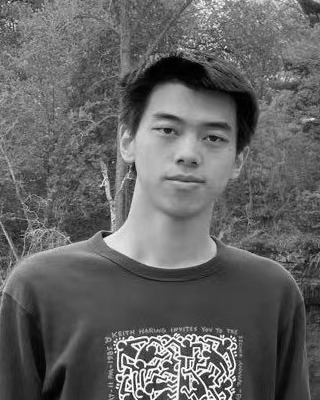}}]{Yuanhao Liu}
earned the B.Eng degree in Electrical Information Engineering from the Chinese University of Hong Kong, Shenzhen, China, in 2019 and the M.S. degree in Electrical and Computer Engineering from University of Minnesota, Minneapolis, USA, in 2020. After graduation, he worked at Huawei Technologies and currently works as a Machine Learning Engineer for a robotics startup in Shanghai, China. His research interests are robotics and machine learning.
\end{IEEEbiography}
\vspace*{-5\baselineskip}

\begin{IEEEbiography}[{\includegraphics[width=1in,height=1.25in,clip,keepaspectratio]{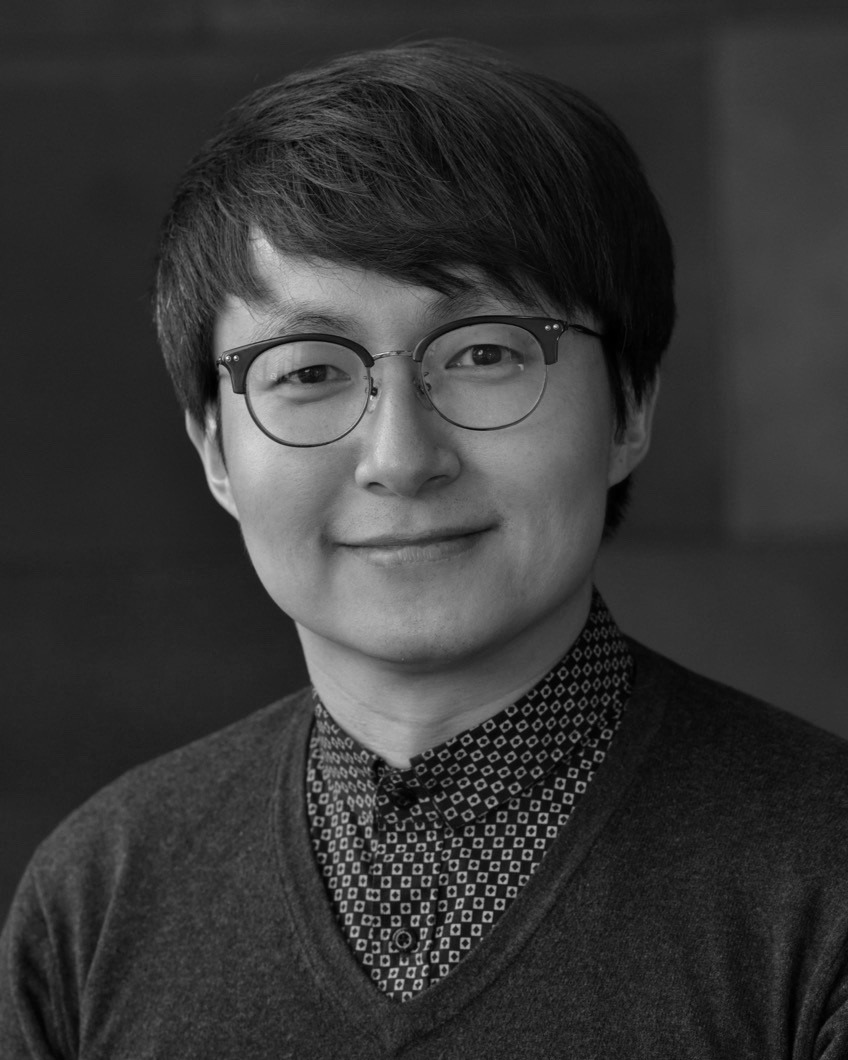}}]{Changhyun Choi} is an Assistant Professor in the Department of Electrical and Computer Engineering at the University of Minnesota. Before joining the UMN, he was a Postdoctoral Associate in the Computer Science \& Artificial Intelligence Lab at Massachusetts Institute of Technology. He obtained a Ph.D. in Robotics at Georgia Institute of Technology. 

Dr. Choi's broad research interests are in visual perception for robotic manipulation, with a focus on deep learning for object grasping and assembly manipulation, soft manipulation, object pose estimation, visual tracking, and active perception. He is the recipient of the NSF CAREER Award, Sony Research Award, Russell J. Penrose Excellence in Teaching Award, and ICRA 2022 Outstanding Student Paper Award.
\end{IEEEbiography}

\clearpage

\end{document}